\documentclass[lettersize,journal]{IEEEtran}
\usepackage{amsmath,amsfonts,amssymb}
\usepackage{algorithmic}
\usepackage{algorithm}
\usepackage{array}
\usepackage[caption=false,font=normalsize,labelfont=sf,textfont=sf]{subfig}
\usepackage{textcomp}
\usepackage{stfloats}
\usepackage{url}
\usepackage{verbatim}
\usepackage{graphicx}

\usepackage{hyperref}
\usepackage{booktabs}     
\usepackage[table]{xcolor}
\usepackage{pifont}
\usepackage{cite}
\usepackage[numbers]{natbib}

\newcommand{\ftcross}{\textcolor{red}{\ding{55}}}
\newcommand{\ftcheck}{\textcolor{teal}{\ding{51}}}

\newcommand{\smalltab}{}

\ifdefined\hidecomments
  \newcommand\minghao[1]{}
  \newcommand\nanda[1]{}
  \newcommand\haoran[1]{}
  \newcommand\charan[1]{}
  \newcommand\kimberly[1]{}
  \newcommand\meet[1]{}
  \newcommand\together[1]{}
\else
  \newcommand\minghao[1]{{\color{blue}Minghao: #1}}
  \newcommand\nanda[1]{{\color{green}Nanda: #1}}
  \newcommand\haoran[1]{{\color{red}Haoran: #1}}
  \newcommand\kimberly[1]{{\color{teal}Kimberly: #1}}
  \newcommand\charan[1]{{\color{orange}Charan: #1}}
  \newcommand\meet[1]{{\color{purple}Meet: #1}}
  \newcommand\together[1]{{\color{brown}All: #1}}
\fi

\begin{document}

\title{D-CIPHER: \underline{D}ynamic \underline{C}ollaborative \underline{I}ntelligent Multi-Agent System with \underline{P}lanner and \underline{H}eterogeneous \underline{E}xecuto\underline{r}s for Offensive Security}
% D-CIPHER: Dynamic Collaborative Intelligent Multi-Agents with Planner and Heterogeneous Executors for Offensive Security

% \title{D-CIPHER: Planner-Executor Multi-Agent Automation Framework For Solving CTF Challenges}

\author{Meet Udeshi*, Minghao Shao*, Haoran Xi*, Nanda Rani, Kimberly Milner, Venkata Sai Charan Putrevu, Brendan Dolan-Gavitt,  Sandeep Kumar Shukla, Prashanth Krishnamurthy, Farshad Khorrami, Ramesh Karri, Muhammad Shafique% <-this % stops a space    
\thanks{*Authors contributed equally to this research.}%
\thanks{M. Udeshi, M. Shao, H. Xi, K. Milner, V.S.C. Putrevu, B. Dolan-Gavitt, P. Krishnamurthy, F. Khorrami, and R. Karri are with the NYU Tandon School of Engineering. M. Shao and M. Shafique are with NYU Abu Dhabi. N. Rani and S.K. Shukla are with the Indian Institute of Technology Kanpur.}% <-this % stops a space
\thanks{This work was supported in part by the NYUAD Center for Artificial Intelligence and Robotics (CAIR), funded by Tamkeen under the NYUAD Research Institute Award CG010, NYUAD Center for Cyber Security (CCS), funded by Tamkeen under the NYUAD Research Institute Award G1104.}}

% \markboth{Transactions on Information Forensics and Security}%
% {Udeshi \MakeLowercase{\textit{et al.}}: D-CIPHER}


\maketitle

\begin{abstract}
Large Language Models (LLMs) have been used in cybersecurity such as autonomous security analysis or penetration testing. Capture the Flag (CTF) challenges serve as benchmarks to assess automated task-planning abilities of LLM agents for cybersecurity. Early attempts to apply LLMs for solving CTF challenges used single-agent systems, where feedback was restricted to a single reasoning-action loop. This approach was inadequate for complex CTF tasks. Inspired by real-world CTF competitions, where teams of experts collaborate, we introduce the D-CIPHER LLM multi-agent framework for collaborative CTF solving. D-CIPHER integrates agents with distinct roles with dynamic feedback loops to enhance reasoning on complex tasks. It introduces the \textit{Planner-Executor agent system}, consisting of a Planner agent for overall problem-solving along with multiple heterogeneous Executor agents for individual tasks, facilitating efficient allocation of responsibilities among the agents. Additionally, D-CIPHER incorporates an \textit{Auto-prompter agent} to improve problem-solving by auto-generating a highly relevant initial prompt. We evaluate D-CIPHER on multiple CTF benchmarks and LLM models via comprehensive studies to highlight the impact of our enhancements. Additionally, we manually map the CTFs in NYU CTF Bench to MITRE ATT\&CK techniques that apply for a comprehensive evaluation of D-CIPHER's offensive security capability. D-CIPHER achieves state-of-the-art performance on three benchmarks: \textbf{22.0\%} on NYU CTF Bench, \textbf{22.5\%} on Cybench, and \textbf{44.0\%} on HackTheBox, which is 2.5\% to 8.5\% better than previous work. D-CIPHER solves 65\% more ATT\&CK techniques compared to previous work, demonstrating stronger offensive capability. D-CIPHER is available at \url{https://github.com/NYU-LLM-CTF/nyuctf_agents} as the \texttt{nyuctf\_multiagent} package. The MITRE ATT\&CK techniques mapping is available at \url{https://github.com/NYU-LLM-CTF/NYU_CTF_Bench} under the \texttt{mitre\_attack\_mapping} folder.
\end{abstract}


\section{Introduction}

% Capture the Flag (CTF) challenges have emerged as a popular platform for testing and honing cybersecurity skills [cite here]. These competitions often involve a series of complex tasks that require a deep understanding of various security topics such as crypto, digital forensics and reverse engineering. These challenges demand not only domain expertise but also strategic problem-solving and adaptability, marking them an ideal test bench for evaluating automated systems. Large language models (LLMs) has shown significant potential in various AI tasks including the application of the LLM agentic systems on cybersecurity including vulnerability detection [cite], bug localization [cite] and automated program repair (APR) [cite]. With these advents of LLMs, there has been a surge of interest in leveraging these models for solving complex tasks, including CTF challenges. Recent work [cite here] indicates that CTF challenges can be a good practice to evaluate the LLM's on its cybersecurity and automated task planning skills as it is a simulation on real world attack scenarios [cite here]. Current developments in LLM-based agentic systems have also shown promising results in automating CTF challenge solutions [cite here], as well as comprehensive benchmarks [cite here] designed specifically for these LLM agents.

\IEEEPARstart{L}{arge} language models (LLMs) have demonstrated remarkable potential in cybersecurity applications such as vulnerability detection 
\cite{lu2024grace, guo2024outside, akuthota2023vulnerability}, bug localization \cite{li2024attention, zhang2024empirical}, and automated program repair \cite{bouzenia2024repairagent, xia2024automated}.
Recent advances in LLMs have led to their application to autonomously perform complex cybersecurity tasks \cite{bhatt2024cyberseceval, wan2024cyberseceval3advancingevaluation}.
Autonomous agents for offensive security are critical to counter the rapidly expanding cyber threats \cite{DARPA-CGC, DARPA-AIxCC, xu2024autopwn}.
Capture the Flag challenges (CTFs)  are suitable for improving cybersecurity skills \cite{chicone2018using, vykopal2020benefits}. CTFs help evaluate LLM proficiency in cybersecurity and automated task planning by simulating real-world offensive security scenarios \cite{tann2023using, yang2023language, shao2024nyu, savin2023battle, pieterse2024friend}, as they contain complex tasks requiring expertise across cryptography, digital forensics, binary exploitation, and reverse engineering.
Autonomous LLM agents are evaluated with jeopardy-style CTFs involving standalone software which after successfully compromising reveals a unique ``flag'' string as a clear indicator of success.
Offensive security capabilities of LLM agents can also be benchmarked using the MITRE ATT\&CK framework \cite{mitre-attack} that offers real-world threat classification \cite{charan2023text, bianou2024pentestmitre}.
% such as a binary to be reverse-engineered, encrypted data to be decrypted, or a web server authentication to be bypassed.
%LLM agents demonstrate potential in autonomously solving these CTF challenges. 
% CTF benchmarks \cite{zhang2024cybenchframeworkevaluatingcybersecurity, shao2024nyu}  advance the  cybersecurity abilities of LLM agents.

\begin{figure}[tpb]
    \centering
    \includegraphics[width=\linewidth]{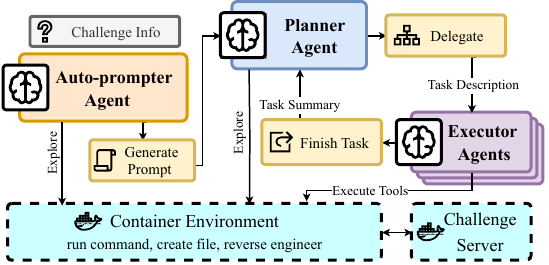}
    \caption{Overview of D-CIPHER. The Auto-prompter, Planner, and heterogeneous Executors all collaborate and interact to solve the CTF.}
    \label{fig:overview}
\end{figure}

% \IEEEpubidadjcol
Most current LLM agents for CTFs operate as single agents handling challenges end-to-end. However, CTFs are complex and require exploration and sequential task execution. Single-agent setups limit feedback to self-reflection, often leading to retries, loss of focus, and hallucinations. In contrast, real-world CTFs are team-based, involving diverse expertise \cite{chang2022capture, cuevas2022observations}, which current frameworks fail to reflect. While multi-agent systems are gaining traction in other fields \cite{dorri2018multi, li2024more, xu2023rewoo}, their use in cybersecurity is still nascent. Offensively, they can automate tasks like pentesting and exploit generation \cite{shen2024pentestagent, chahal2023ai}; defensively, they aid in bug discovery and repair \cite{le2012systematic}. Motivated by this, we propose a multi-agent LLM framework that assigns distinct roles to agents, enabling dynamic and collaborative problem-solving.

%Current LLM agents for CTFs operate as single agents that handle the challenge from start to finish. CTFs are complex tasks that require significant exploration and proper execution of a sequential tasks. Single agents typically restrict feedback to self-reflection within the LLM’s context, going through multiple retries for a single task. This causes loss of focus on the overall problem and hallucinations about task details. On the other hand, real-world CTF competitions are approached collaboratively by teams \cite{chang2022capture, cuevas2022observations} of members with diverse expertise. Current agent frameworks cannot capture the collaborative nature of CTF competitions. Multi-agent systems have gained prominence across different domains \cite{dorri2018multi, li2024more, xu2023rewoo} but their application to cybersecurity remains unexplored.  In offensive security, they can automate tasks like penetration testing, vulnerability detection, or incident response by analyzing attack patterns, attempting multiple exploits, and producing detailed reports \cite{shen2024pentestagent, chahal2023ai}. For defensive applications, multi-agent systems can automate bug discovery, bug replication, and program repair \cite{le2012systematic}. Inspired by this, we introduce an LLM multi-agent framework that divides responsibilities among multiple agents to facilitate dynamic interactions and allow collaborative problem-solving.

We present D-CIPHER, a novel LLM multi-agent framework to autonomously solve CTFs via collaboration of multiple LLM agents. 
D-CIPHER introduces two mechanisms for enhanced interaction and dynamic feedback between  agents:
first, the \textit{Planner-Executor agent system} that involves a Planner to solve the CTF end to end, and multiple heterogeneous Executor agents to complete single tasks assigned by the Planner;
and second, the \textit{Auto-prompter agent} to explore the CTF environment and generate an initial prompt for the main system.
Dividing responsibilities between planner and executors allows each agent to maintain focus for long complex tasks, and reduces hallucinations.
Auto-prompting is a prompt engineering technique to improve LLM performance by generating dynamic task-specific prompts as opposed to human-written hard-coded prompt templates. D-CIPHER incorporates auto-prompting as a separate agent to produce a highly-relevant initial prompt to kick-start the main system.
Additionally, we are the first work to evaluate LLM agents using the MITRE ATT\&CK framework \cite{mitre-attack}. We augment the NYU CTF Bench \cite{shao2024nyu} with a mapping of ATT\&CK techniques to evaluate D-CIPHER and related LLM agents by the techniques they employ for the CTFs, offering a comprehensive overview of the agent's offensive security capability. 

Figure~\ref{fig:overview} shows an overview of D-CIPHER. All agents access a shared container environment to run shell commands and interact with the CTF server.
The Auto-prompter starts the process by exploring the environment and generating a prompt for the Planner. The Planner also explores for a few rounds, after which it creates a plan and delegates tasks to the Executors. Each delegated task initiates a new Executor with a new conversation history, allowing for heterogeneous execution and greater focus on single tasks. After completing the task, the Executor returns a task summary which the Planner may use to update the plan and delegate further tasks. The Planner-Executor loop continues until the challenge is solved, or some terminal conditions are met.
This collaborative design allows D-CIPHER to tackle complex CTFs, improving performance to achieve state-of-the-art accuracy on CTF benchmarks.

We evaluate D-CIPHER seven LLM models, via it's accuracy on three benchmarks and it's performance in solving MITRE ATT\&CK technqiues. Our results demonstrate that the multi-agent approach not only improves problem-solving, but also enhances robustness by mitigating errors and dynamically adapting strategies during runtime. We perform ablation studies and comparison with related works to further illustrate D-CIPHER's ability to outperform single-agent systems. Performance on the MITRE ATT\&CK techniques additionally reveals D-CIPHER's superior offensive security capability.
The contributions of this work are as follows: 
\begin{enumerate}
    \item \emph{D-CIPHER}, a novel LLM multi-agent framework that leverages specialized agents with distinct roles to enable agent collaboration for autonomous problem-solving
    \item A novel \emph{Planner-Executor system} with a Planner and multiple Executors to divide responsibilities and enhance long-term focus for complex problems
    \item A novel \emph{Auto-prompter agent} that improves auto-prompting with an agent setup
    \item Augmenting the NYU CTF Bench by mapping  MITRE ATT\&CK techniques and elaborating D-CIPHER's offensive security capability
    \item A comprehensive study on how multi-agent collaboration between agents enhances problem-solving on CTFs
\end{enumerate}

% The paper is structured as follows: 
% Section~\ref{sec:background} presents background on LLM agents and auto-prompting,
% Section~\ref{sec:related_work} compares related work,
% Section~\ref{sec:implementation} describes D-CIPHER's implementation,
% Section~\ref{sec:experiment_setup} provides the experiment setup,
% Section~\ref{sec:results} presents the results along with comprehensive studies of D-CIPHER's performance,
% Section~\ref{sec:discussion} discusses two case studies, limitations, and ethics,
% and Section~\ref{sec:conclusion} concludes and proposes future work.

The paper is structured as follows: 
Section~\ref{sec:background_related} provides background and reviews related work, Section~\ref{sec:implementation} describes D-CIPHER's implementation, Section~\ref{sec:experiment_setup} outlines the experimental setup, Section~\ref{sec:results} presents the results, Section~\ref{sec:discussion} discusses common failures and ethics, and Section~\ref{sec:conclusion} concludes the paper and proposes directions for future work.

\section{Related Work} \label{sec:background_related}

%Instead of prompting the LLM in a question-answer or conversational manner, 
Autonomous frameworks create a feedback loop to allow the LLM to perform tasks and operate as autonomous agents.
% \cite{yang2023intercode, yang2024sweagent, abramovich2024enigma, shao2024nyu}.
LLMs are supporting function (or tool) calling where actions can be provided that the LLM may choose to ``call'' as a function.
Many ``tools'' can be provided such as a command line, web search, file editing, and code execution \cite{wang2024surveyllmagents}.
To help LLMs on long-horizon tasks, plan-and-solve prompting \cite{wang2023planandsolve} enhances long-term focus via a planning phase before iterative execution to tackle complex tasks \cite{turtayev2024hacking}.
ReAct (reasoning + action) \cite{yao2022react} combines step-by-step reasoning with action.
% ReWOO (Reasoning without Observation) \cite{xu2023rewoo} separates the reasoning process from tool outputs and observations.
The prompting methods in these agents involve static hard-coded templates where environment and task information is filled in.
These often fail to adapt to different problems.
Auto-prompting~\cite{shin-etal-2020-autoprompt, zhou-etal-2023-revisiting, zhang2023automatic} allows the LLM itself to generate a highly-relevant prompt, invoking factual responses and reducing hallucinations.
D-CIPHER incorporates auto-prompting as a separate agent that can explore the environment and generate a better prompt.
Expanding on single LLM agents, Multi-agent systems enhance problem-solving by collaboration between specialized agents, working on different aspects of complex tasks \cite{guo2024largelanguagemodelbased}.
Multi-agent systems are effective in cybersecurity applications such as insider threat detection \cite{song2024audit}, incident response \cite{liu2024multi}, and improving code safety \cite{nunez2024autosafecoder}.

% \begin{table}[tpb]
%     \centering
%     \small
   
%     \begin{tabular}{p{0.43\linewidth}ccccc}
%     \toprule
%          \textbf{Work} & \rotatebox{90}{\textbf{\# CTFs}} & \rotatebox{90}{\textbf{Tool use}}  & \rotatebox{90}{\textbf{Autonomous}} & \rotatebox{90}{\textbf{Multi-agent}} &\rotatebox{90}{\textbf{Auto-prompt}} \\
%     \cmidrule{2-6}
%      % \textbf{Study} & \textbf{Dynamic} & \textbf{Used} & \textbf{Multi-} & \textbf{Automatic} & \textbf{Tool} & \textbf{\# of} \\
%          \citet{tann2023using} &  $7$  & \ftcross & \ftcross & \ftcross & \ftcross  \\
%          % \citet{shao2024empirical} & $26$ & \ftcheck & \ftcheck & \ftcross & \ftcross  \\
%          InterCode-CTF \cite{yang2023language} & $100$  & \ftcheck & \ftcheck & \ftcross & \ftcross   \\
%          NYU CTF Bench \cite{shao2024nyu} & $200$  & \ftcheck & \ftcheck & \ftcross & \ftcross \\
%           \citet{turtayev2024hacking} & $100$ & \ftcheck & \ftcheck & \ftcross & \ftcross\\
%          Cybench \cite{zhang2024cybenchframeworkevaluatingcybersecurity} & $40$  & \ftcheck & \ftcheck & \ftcross & \ftcross \\
%          EnIGMA \cite{abramovich2024enigma} & $350$ & \ftcheck & \ftcheck & \ftcross & \ftcross\\
%          HackSynth \cite{muzsai2024hacksynth} & $200$  & \ftcheck & \ftcheck & \ftcheck & \ftcross \\
%          \textbf{D-CIPHER (ours)} & $290$ & \ftcheck & \ftcheck & \ftcheck & \ftcheck \\
%     \bottomrule
%     \end{tabular}
%      \caption{Comparison of LLM agents for CTFs.}
%     \label{tab:related_work_comparison}
% \end{table}

\begin{table}[htpb]
    \centering
    \caption{Feature comparison of CTF solving agents.}
    \label{tab:related_work_comparison}
    \begin{tabular}{lcccccc}
    \toprule
         \textbf{Study} & \rotatebox{90}{\textbf{\# CTFs}} & \rotatebox{90}{\textbf{Open bench}} & \rotatebox{90}{\textbf{Tool use}}  & \rotatebox{90}{\textbf{Autonomous}} & \rotatebox{90}{\textbf{Multi-agent}} &\rotatebox{90}{\textbf{Auto-prompt}} \\
    \cmidrule{2-7}
     % \textbf{Study} & \textbf{Dynamic} & \textbf{Used} & \textbf{Multi-} & \textbf{Automatic} & \textbf{Tool} & \textbf{\# of} \\
         Tann et al. \cite{tann2023using} &  $7$ & \ftcross & \ftcross & \ftcross & \ftcross & \ftcross  \\
         Shao et al. \cite{shao2024empirical} & $26$ & \ftcross & \ftcheck & \ftcheck & \ftcross & \ftcross  \\
         InterCode-CTF\cite{yang2023language} & $100$ & \ftcheck & \ftcheck & \ftcheck & \ftcross & \ftcross   \\
         NYU CTF Bench \cite{shao2024nyu} & $200$ & \ftcheck & \ftcheck & \ftcheck & \ftcross & \ftcross \\
         Turtayev et al. \cite{turtayev2024hacking} & $100$ & \ftcheck & \ftcheck & \ftcheck & \ftcross & \ftcross\\
         Cybench \cite{zhang2024cybenchframeworkevaluatingcybersecurity} & $40$ & \ftcheck & \ftcheck & \ftcheck & \ftcross & \ftcross \\
         EnIGMA \cite{abramovich2024enigma} & $350$ & \ftcheck & \ftcheck & \ftcheck & \ftcross & \ftcross\\
         HackSynth \cite{muzsai2024hacksynth} & $200$ & \ftcheck & \ftcheck & \ftcheck & \ftcheck & \ftcross \\
         \textbf{D-CIPHER (ours)} & $290$ & \ftcheck & \ftcheck & \ftcheck & \ftcheck & \ftcheck \\
    \bottomrule
    \end{tabular}
\end{table}

Recent works build LLM agents targeted towards CTFs.
Table~\ref{tab:related_work_comparison} shows a feature comparison of D-CIPHER with related works on LLM agents for autonomous CTF solving.
The InterCode-CTF agent~\cite{yang2023intercode} reveals that LLM agents demonstrate basic cybersecurity skills but struggle with more complex tasks.
The NYU CTF baseline agent~\cite{shao2024empirical} integrates external tools and shows  improved potential of tool-assisted LLMs to solve CTFs, however the agent exhausts LLM context length when command output history grows. InterCode-CTF manages this by truncating the history to the last few iterations. Even so, agents face issues with longer tasks.
Agents perform better with a focused set of tools with well-defined interfaces~\cite{yang2024sweagent}.
EnIGMA~\cite{abramovich2024enigma} agent incorporates interactive tools, in-context learning, and LLM summarizer for context management to achieve state-of-the-art results. 
%EnIGMA also uses an LLM summarizer that summarizes the command outputs for the main agent for context management.
HackSynth~\cite{muzsai2024hacksynth} uses iterative planning and feedback summarization stages which helps to finish multiple tasks and improves overall problem solving.
Similarly, Cybench~\cite{zhang2024cybenchframeworkevaluatingcybersecurity} introduces a benchmark of 40 CTFs augmented with step-by-step tasks, focusing LLM agents on each smaller task.
\citet{turtayev2024hacking} expand on InterCode-CTF by implementing plan-and-solve prompting, significantly improving on InterCode-CTF benchmark. 
These works highlight that LLM agents excel at implementing code and executing commands to accomplish small concrete tasks when provided with dynamic feedback and task-specific toolsets. While these works involved multiple LLMs with different tasks such as planning and summarizing along-side a main agent, D-CIPHER is the first work to formulate a multi-agent system for CTFs with division of responsibilities and well-defined interactions for dynamic feedback.

\section{D-CIPHER Implementation} \label{sec:implementation}

\begin{figure*}[htpb]
    \centering
    \includegraphics[width=\linewidth]{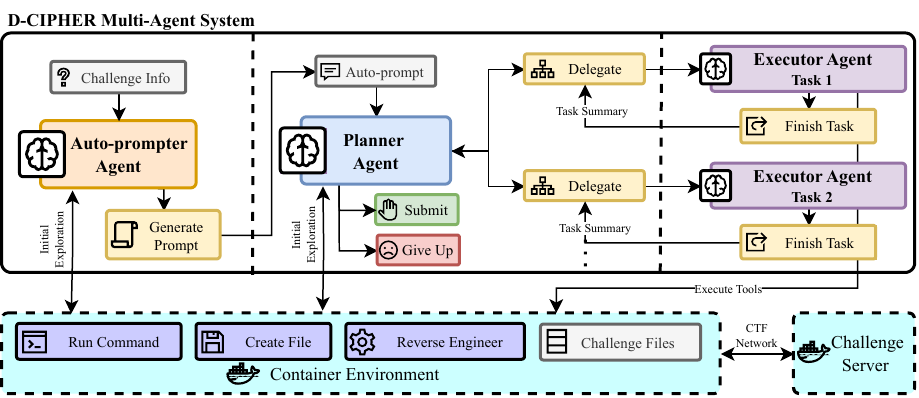}
    \caption{Workflow of the D-CIPHER multi-agent system. Execution starts with the Auto-prompter which explores the CTF and produces a dynamic, relevant prompt. The Planner proceeds with exploration and delegates specific tasks to the Executors. Each Executor starts with a fresh conversation history to focus on the delegated task, while the Planner maintains overall context and drives the problem solving.}
    \label{fig:workflow}
\end{figure*}

The D-CIPHER framework introduces a collaborative LLM multi-agent system. Each individual agent is based on the NYU CTF baseline agent \cite{shao2024nyu} with upgraded prompts that describe tasks and additional interaction tools for a multi-agent collaborative context. We use function calling features of current LLMs to prompt for agent actions.  The system has three agents:
(1) the \emph{Planner agent} generates the overall plan to solve the CTF challenge, delegating specific tasks to Executors, and revising the plan based on their feedback;
(2) the \emph{Executor agent} performs the task delegated by the Planner and returns a summary;
and (3) the \emph{Auto-prompter agent} generates a dynamic prompt based on it's exploration of the CTF.
Figure~\ref{fig:workflow} shows D-CIPHER's workflow.

\subsection{Context Management}
Each agent maintains a conversation history of LLM inputs and outputs.
An LLM agent's context contains: (1) the system prompt that sets the agent's role and provides actions, (2) the initial prompt that describes the environment and the task (e.g., CTF challenge or delegated task); and (3) the conversation history of agent actions and observations.
Following the ReAct strategy, we prompt the LLM to reason and produce an action.
We utilize the function calling features of current LLMs to produce actions, so we do not define a structured format of our own, but instead rely on the LLM provider's API to parse the actions correctly.
At every iteration, the conversation history is sent to the LLM and it generates a message containing the reason and action. Observations from executing the actions are appended to the conversation history.
The generated reason, action, and corresponding observation constitutes a  ``round'' of conversation.
The agent continue these rounds until the task is complete or the context is full.
To avoid filling up the context, we truncate observations to 25,000 characters. We also optionally truncate actions and observations in all but the last few rounds while keeping the reasoning intact, similar to \citet{abramovich2024enigma}. %The second technique is only applied to the Executor, as it has been observed to help execute long tasks.

\subsection{Environment and Tools}

All agents interact with the same Linux container environment containing the CTF files and providing network access to the CTF server and the internet to install new packages.
The agents have access to the following tools:
\texttt{RunCommand} to execute shell commands;
\texttt{CreateFile} to create a file;
\texttt{Disassemble} and \texttt{Decompile} to trigger Ghidra\footnote{Ghidra is a popular reverse engineering tool. \url{https://ghidra-sre.org/}.} to reverse engineer a binary;
\texttt{SubmitFlag} to submit a CTF flag to solve the challenge;
and, \texttt{Giveup} to giveup solving.
Unlike EnIGMA \cite{abramovich2024enigma}, we do not implement advanced interfaces or interactive tools.
The specialized reverse engineering tools offer the agents access to Ghidra which does not provide a direct command line interface.
%Specialized tools for other categories, like \texttt{RsaCtfTool} for cryptography or \texttt{nikto} for web, are mentioned in the category-specific initial prompt because they can be run from the command line via \texttt{RunCommand}.
We also provide special actions for interaction between agents:
\texttt{GeneratePrompt} for the Auto-prompter to generate a prompt;
\texttt{Delegate} for the Planner to delegate tasks;
and, \texttt{FinishTask} for the Executor to terminate and return a task summary.

\subsection{Workflow}

% \begin{figure*}[tbp]
%     \centering
%     \includegraphics[width=\linewidth, trim=30 30 30 30]{figures/1black0white_example.pdf}
%     \caption{Interaction between Planner and Executors for the \texttt{1black0white} forensics challenge.}
%     \label{fig:planner_conv}
% \end{figure*}

% \begin{figure}[htbp]
%     \centering
%     \includegraphics[width=\linewidth]{figures/autoprompt_small_vert.pdf}
%     \caption{Auto-prompter generated prompt vs. hard-coded template for the \emph{1black0white}  CTF.}
%     \label{fig:autoprompt_conv}
% \end{figure} 

\begin{figure}[htpb]
    \centering
    \includegraphics[width=\linewidth]{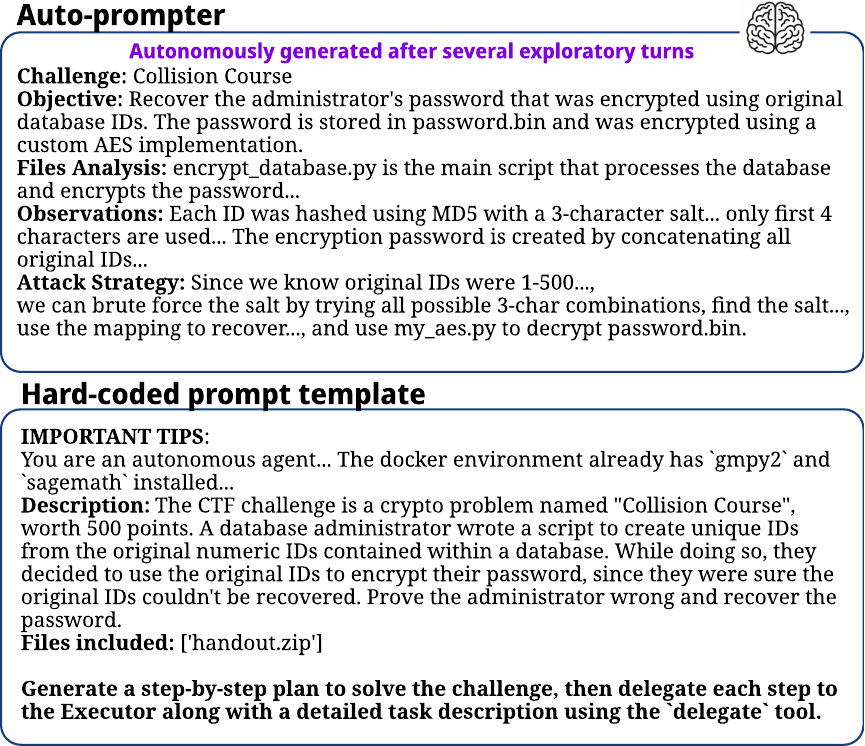}
    \caption{Auto-prompter generated prompt vs. hard-coded template for the \emph{collision\_course} CTF. Auto-prompter's dynamic prompt captures the approach tailored for this CTF.}
    \label{fig:autoprompt_conv}
\end{figure}

\begin{figure}[htpb]
    \centering
    \includegraphics[width=\linewidth]{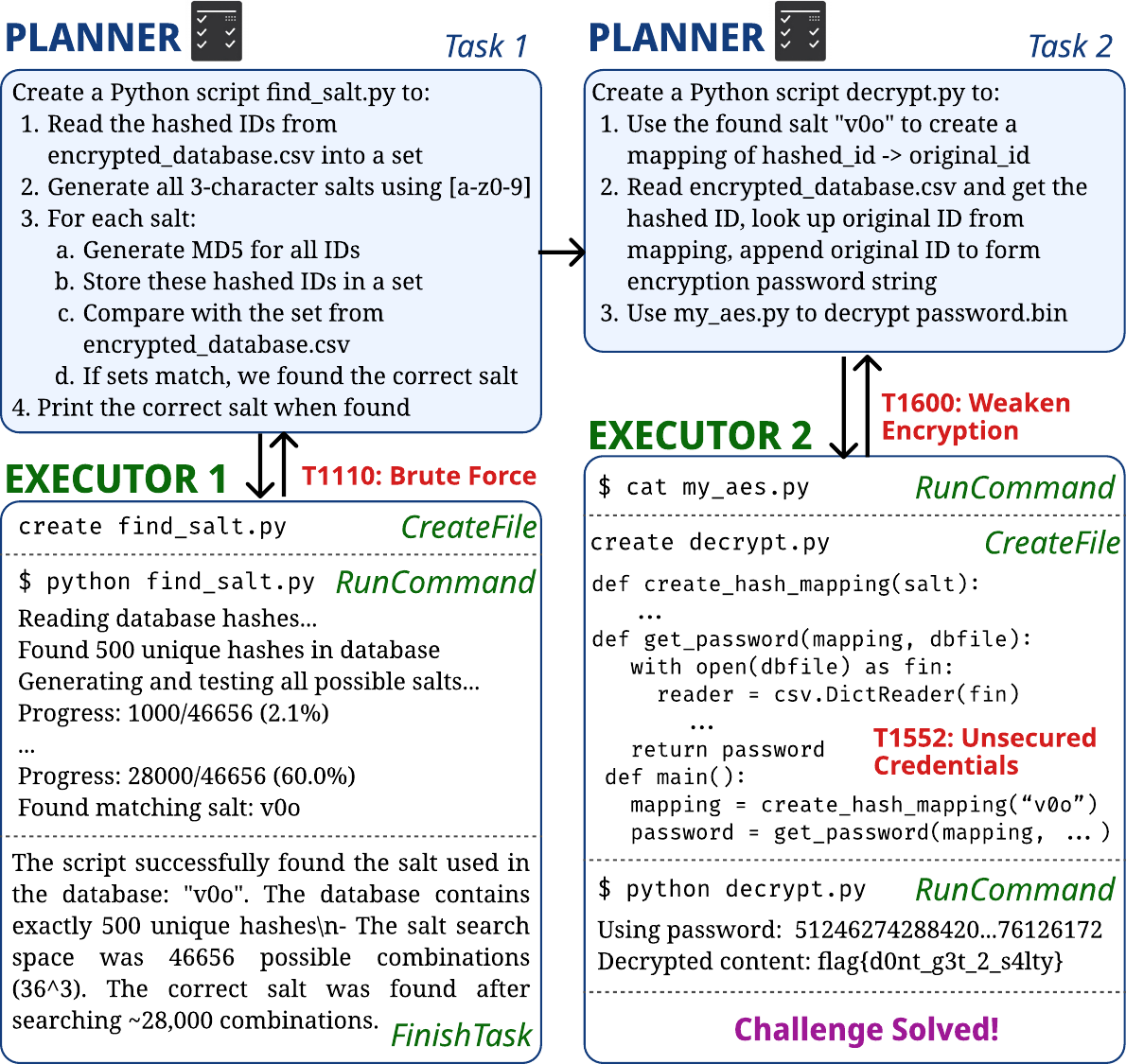}
    \caption{Planner and Executors interact for the \emph{collision\_course} cryptography CTF. Planner drives the problem solving, while each Executor focuses on delegated tasks and implements specific MITRE ATT\&CKs.}
    \label{fig:planner_conv}
\end{figure}

\subsubsection{Auto-prompter} 
The Auto-prompter agent is provided the CTF information in it's initial prompt.
Instead of directly rewriting this prompt like typical auto-prompting applications,
the agent first interacts with the CTF environment for a few rounds by running commands to read available files, execute the CTF binary, or access the CTF server.
Based on this exploration, it generates a prompt tailored for this CTF by calling \texttt{GeneratePrompt}.
Figure~\ref{fig:autoprompt_conv} shows an example auto-prompt and the hard-coded prompt for the \emph{collision\_course} cryptography CTF from NYU CTF Bench.
Along with a relevant description, the Auto-prompter proposes a viable approach based on it's exploration.
The hard-coded templates provide generic directions that cannot be tailored to each CTF.

\subsubsection{Planner-Executor system} 

The Planner is initiated with the generated prompt and also explores the CTF for a few turns.
It is provided with \texttt{RunCommand} but not \texttt{CreateFile}, \texttt{Disassemble}, or \texttt{Decompile}, allowing exploration, but dissuading it from trying to solve the CTF by itself.
It comes up with a step-by-step plan and delegates tasks to an Executor by calling \texttt{Delegate}.
The Executor is initiated with the task details and performs the task by running commands and creating scripts, after which it calls \texttt{FinishTask} with a task execution summary and results.
The summary is returned to the Planner as an observation, using which the Planner continues to revise its plan and delegate further tasks.
For each \texttt{Delegate} call, the framework initiates a new Executor with a new conversation history.
Effectively, D-CIPHER runs multiple heterogeneous Executors to solve one challenge.
Each Executor focuses on it's own task, while the Planner only sees the task summary, allowing for efficient context management.
Figure~\ref{fig:planner_conv} shows an example of the Planner solving the \texttt{collision\_course} CTF.
Based on the Auto-prompter's suggested approach and it's own exploration, the Planner starts by delegating the task of cracking the hash salt using brute force.
The first Executor successfully implements the brute force attack, correctly employing the T1110 (Brute Force) ATT\&CK technique, and returns the task summary along with the hash salt to the Planner.
The Planner then reasons and delegates the next step to use the salt and decrypt the password.
The second Executor implements the decryption script, employing the two techniques T1600 (Weaken Encryption) and T1552 (Unsecured Credentials).
Successfully executing the script reveals the flag and solves the CTF.
This example shows how the Planner focuses on the entire CTF while each Executor focuses on single tasks.
The workflow ensures continuous interaction between the Planner and Executors such that they collaborate to enhance problem-solving.
Enhanced focus on single tasks also improves D-CIPHER's capabilities on MITRE ATT\&CK techniques.

% The Planner starts with two simple exploratory tasks of examining the CTF file, to which the Executor returns a summary of file contents. Based on that, the Planner reasons and delegates the next steps to convert the file and parse the flag.
% This example shows how the Planner focuses on the entire CTF while each Executor focuses on single tasks.
% The workflow ensures continuous interaction between the Planner and Executors such that they collaborate to enhance problem-solving.

For each agent, we define a maximum number of conversation rounds after which the agent is stopped. We also set a maximum cost limit of all agents.
Among all agents, only the Planner has access to \texttt{SubmitFlag} or \texttt{Giveup} tools, making it the central agent of D-CIPHER.
D-CIPHER terminates when \texttt{SubmitFlag} is called with the correct flag, \texttt{Giveup} is called, the Planner exhausts its rounds, or the cost limit is reached.
If a wrong flag is submitted, a negative response is returned and solving may continue.
The Auto-prompter and Executor sometimes exhaust their conversation rounds only running commands and fail to produce an output for the Planner by calling \texttt{GeneratePrompt} or \texttt{FinishTask}.
In that case, we prompt the agents one last time and insist on producing an output.
If the Auto-prompter fails, a hard-coded prompt is used. If the Executor fails, a hard-coded warning is returned. To improve focus and avoid exceeding the LLM context, we truncate the Executor’s conversation history to include only recent actions and observations.
%If the Auto-prompter still fails, th hard-coded prompt is used. If the Executor still fails, a hard-coded warning is returned. Only for the Executor, we truncate the conversation history to contain last few actions and observations, which prevents filling up the LLM context and improves focus.

%If the Auto-prompter still does not call \texttt{GeneratePrompt}, the hard-coded prompt is used.
%If the Executor still does not call \texttt{FinishTask}, a hard-coded warning is returned.

% D-CIPHER's special interaction functions allow versatility to configure different types of multi-agent systems.
% For example, a simpler system can have the Auto-prompter generate a prompt for a single Executor to solve the challenge end-to-end.
% This is implemented for the ablation study in Section~\ref{sec:ablation}. 
% Such configurations demonstrate the framework's flexibility to build systems for different problems.

\section{Experiment Setup} \label{sec:experiment_setup}

Each run of D-CIPHER attempts one CTF challenge.
D-CIPHER is configured as follows:
a total cost limit of \$3,
a temperature of 1.0 for each LLM,
5 max rounds for the Auto-prompter,
30 max rounds for the Planner,
100 max rounds for each Executor,
and each Executor's conversation history is truncated to last 5 actions and observations.

\subsection{Benchmarks}

We evaluate D-CIPHER on NYU CTF Bench \cite{shao2024nyu}, Cybench~\cite{zhang2024cybenchframeworkevaluatingcybersecurity}, and HackTheBox \cite{hackthebox}.
As shown in Table~\ref{table:benchmarks}, these benchmarks have 290 CTFs spanning six categories: cryptography (crypto), forensics, binary exploitation (pwn), reverse engineering (rev), web, and miscellaneous (misc).
We perform ablation studies on NYU CTF Bench.
During development, we use the development set of 55 CTFs introduced in \citet{abramovich2024enigma}.
We use the unguided mode of Cybench that does not include additional subtask information with the ``hard prompt'' that does not contain extra hints.

\begin{table}[htpb]
\centering
\caption{Benchmarks for evaluating D-CIPHER.}
\label{table:benchmarks}
\begin{tabular}{lccccccc}
\toprule
                   & crypto      & foren   & pwn         & rev         & web        & misc         & \textbf{Total} \\ 
\cmidrule{2-8}
NYU CTF            & 53          & 15          & 38          & 51          & 19          & 24          & \textbf{200}   \\
Cybench            & 16          & 4           & 2           & 6           & 8           & 4           & \textbf{40}    \\ 
HackTheBox         & 30          & 0           & 0           & 20          & 0           & 0           & \textbf{50}    \\
\midrule
\textbf{Total}     & \textbf{99} & \textbf{19} & \textbf{40} & \textbf{77} & \textbf{27} & \textbf{28} & \textbf{290}   \\ 
\bottomrule
\end{tabular}
\end{table}

\subsection{LLM Selection}

% We test multiple LLMs. %with D-CIPHER for each of the Planner, Executor, and Auto-prompter agents. 
For our experiments, we use the same LLM for all three agents.
We access LLMs via their APIs. Open-source LLaMa models are accessed via the Together AI platform\footnote{\url{https://www.together.ai}}.
We use the following LLMs:
Claude 3.5 Sonnet (\emph{claude-3-5-sonnet-20241022}),
GPT 4 Turbo (\emph{gpt-4-turbo-2024-04-09}),
GPT 4o (\emph{gpt-4o-2024-11-20}),
LLaMa 3.1 405B (\emph{meta-llama/Meta-Llama-3.1-405B-Instruct-Turbo}),
and Gemini 1.5 Flash (\emph{gemini-1.5-flash}).
%We try with different LLMs for Planner and Executor in Appendix~\meet{fix ref}.%\ref{sec:combiningllms}.

D-CIPHER supports different LLMs for each agent. We explore configurations that pair stronger models for the Planner with weaker models for the Executor. The weaker LLMs used include Claude 3.5 Haiku (\emph{claude-3-5-haiku-20241022}), GPT-4o Mini (\emph{gpt-4o-mini-2024-07-18}), LLaMa 3.3 70B (\emph{meta-llama/Llama-3.3-70B-Instruct-Turbo}), and Gemini 1.5 Flash 8B (\emph{gemini-1.5-flash-8b}).
%D-CIPHER offers freedom to use different LLMs for each agent, and we experiment by combining stronger models for the Planner with weaker models for the Executor. We use the following weaker LLMs: Claude 3.5 Haiku (\emph{claude-3-5-haiku-20241022}),
GPT 4o Mini (\emph{gpt-4o-mini-2024-07-18}), LLaMa 3.3 70B (\emph{meta-llama/Llama-3.3-70B-Instruct-Turbo}),
and Gemini 1.5 Flash 8B (\emph{gemini-1.5-flash-8b}).

% Models such as GPT-4 Turbo (gpt-4-turbo-2024-04-09), GPT-4o (gpt-4o-2024-11-20), Claude 3.5 Sonnet (claude-3-5-sonnet-20241022), LLaMA 3.1 (405B), and Google's Gemini 1.5 Flash are employed in both planner and executor roles. Additionally, as part of our ablation study, we assessed D-CIPHER's performance using strong-weak model combinations across various agent roles. To achieve this, the stronger models were paired with complementary executor models, including Claude 3.5 Haiku (claude-3-5-haiku-20241022), GPT-4o Mini (gpt-4o-mini-2024-07-18), LLaMA 3.1 (70B), and Gemini Flash 8B. This configuration was designed to analyze the effectiveness of the planner-executor combinations across different benchmarks. For API access, we use the OpenAI API for GPT models, Anthropic Inference API \cite{anthropicapi} for Claude models, the Google API for Gemini, and the Together API \cite{togetherai} for LLaMA models. All models are configured with a default temperature of 1.0 to ensure consistency in performance evaluation. 

\subsection{Evaluation Metrics}

The primary evaluation metric is percentage of CTFs successfully solved (\textit{\% solved}). 
A CTF is solved when the correct flag is submitted by the Planner or if the correct flag is observed in the agent conversation. The latter prevents failures where the Auto-prompter or Executors find the flag but do not tell the Planner, as only the Planner can submit a flag.
False positives are highly unlikely because flags are unique strings with specific formats such as \texttt{flag\{...\}}.
% This approach mimics real-world CTFs where participants may submit a flag multiple times and receive instant confirmation.
We also measure the average cost of solved CTFs (\textit{\$ cost}).
The total cost of one CTF is the US dollar cost of all LLM API calls across agents. 
The API cost is indicative of the computational resources required to deploy LLMs, so this metric estimates computational resources for solved CTFs.

\subsection{MITRE ATT\&CK Classification} \label{sec:mitre_label}

The MITRE ATT\&CK framework \cite{mitre-attack} is a popular taxonomy of offensive security tactics, techniques and procedures used to classify cyber attacks \cite{charan2023text}. CTF challenges emulate real-world cyberattack scenarios and we can attribute a CTF to a list of ATT\&CK techniques that must be employed to solve that CTF. For all the CTFs that an agent solves, we aggregate each CTF's mapped techniques that the agent successfully employs to comprehensively benchmark its offensive security capability using the ATT\&CK framework.
 
To perform this analysis, we manually labeled the 200 CTFs in NYU CTF Bench with ATT\&CK enterprise techniques as a part of this work. We performed the labeling based on the CTF description, solution writeups and scripts, and manual interaction with the CTF. We then mapped the ATT\&CK techniques that must be employed to solve the CTF. Some CTFs only test specific skills and do not involve any attack, especially in cryptography, reverse engineering and miscellaneous categories, so 83 of the 200 CTFs have no techniques that apply.
On the remaining 117 CTFs, we mapped 211 instances of 45 unique techniques. 
Table~\ref{tab:mitre_analysis} in Section~\ref{sec:mitre_result} shows the list of techniques and number of CTFs that they apply to.
The frequently occurring techniques such as T1600 (Weaken Encryption) and T1552 (Unsecured Credentials) apply to many cryptography CTFs, while T1203 (Exploitation for Client Execution) and T1574 (Hijack Execution Flow) apply to binary exploitation CTFs. Performance on these techniques reflects the category-wise accuracy and  offers granular insights into offensive capability. %, especially when viewed in combination with performance on infrequent techniques.
% \meet{Have we?} \minghao{I think it is mainly about some specific tools used or some papers introduced with the challenge (one example is crypto lottery in 2023q which get inspired by a paper)} We have included the labeling data in supplementary material.
% We did not label sub-techniques, instead labeled the encompassing technique where a sub-technique applied.

\section{Results} \label{sec:results}

\begin{table*}[tpb!]
    \centering
    \caption{Performance across different models and configurations on NYU CTF Bench, Cybench, and HackTheBox benchmarks.} %. Bold values indicate the highest scores within each category.}
    \label{tab:default_experiment}
    \begin{tabular}{lcccccccccccc}
    \toprule
    & \multicolumn{8}{c}{\textbf{NYU CTF Bench}} & \multicolumn{2}{c}{\textbf{Cybench}} & \multicolumn{2}{c}{\textbf{HackTheBox}} \\
    & \% solved & \$ cost & crypto & forensics & pwn & rev & web & misc & \% solved & \$ cost & \% solved & \$ cost  \\
    \cmidrule(lr){2-9} \cmidrule(lr){10-11} \cmidrule(lr){12-13}
    \textbf{NYU CTF baseline} & & & & & & & & & & & & \\
    \smalltab Claude 3.5 Sonnet & 13.0 & -- & 7.7 & \textbf{20.0} & 7.7 & 21.6 & 5.3 & 16.7 & 15.0 & -- & 38.0 &-- \\
    \smalltab GPT 4o & 6.0 & -- & 3.8 & 0.0 & 5.1 & 9.8 & 0.0 & 12.5 & 12.5 & -- & 16.0 & -- \\
    \smalltab GPT 4 Turbo & 6.0 & -- & 1.9 & 0.0 & 5.1 & 9.8 & 0.0 & 16.7 &  12.5 & -- & 10.0 &-- \\
    % \textbf{Cybench baseline} & & & & & & & & & & & & \\
    % \smalltab Claude 3.5 Sonnet & -- & -- & -- & -- & -- & -- & -- & -- & 17.5 & -- & -- & -- \\
    % \smalltab GPT 4o & -- & -- & -- & -- & -- & -- & -- & -- & 12.5 & -- & -- &-- \\
%    \smalltab LLaMa 3.1 405B & -- & -- & -- & -- & -- & -- & -- & -- & 7.5 & -- & -- &-- \\
    \textbf{EnIGMA} & & & & & & & & & & & & \\
    \smalltab Claude 3.5 Sonnet & 13.5 & 0.35 & 7.7 & \textbf{20.0} & 18.0 & 17.7  & 0.0 & 16.7 & 20.0 & 0.91 & 26.0 & 0.53 \\
    \smalltab GPT 4o & 9.5 & 0.62 & 3.9 & 13.3 & 7.7 & 13.7 & 5.3 & 16.7 & 12.5 & 0.61 & 16.3 & 1.71 \\
    \smalltab GPT 4 Turbo & 7.0 & 0.79 & 1.9 & 13.3 & 5.1 & 9.8 & 0.0 & 16.7 & 17.5 & 1.60 & 18.4 & 1.35 \\
 %   \smalltab LLaMa 3.1 405B & -- & -- & -- & -- & -- & -- & -- & -- & 10.0 & 0.42 & -- & -- \\
    \textbf{D-CIPHER} & & & & & & & & & & & & \\
    \smalltab Claude 3.5 Sonnet & 19.0 & 0.52 & \textbf{15.4} & \textbf{20.0} & 12.8 & \textbf{29.4} & 5.3 & \textbf{25.0} & \textbf{22.5} & 0.30 & \textbf{44.0} & 0.49 \\
    \smalltab GPT 4o & 10.5 & 0.22 & 5.8 & 13.3 & 7.7 & 13.7 & \textbf{10.5} & 16.7 &  12.5 & 0.08 & 16.0 & 0.16 \\
    \smalltab GPT 4 Turbo & 6.5 & 0.46 & 1.9 & 13.3 & 5.1 & 7.8 & 5.3 & 12.5 &  -- & -- & -- &-- \\
    \smalltab LLaMa 3.1 405B & 3.0 & 0.01 & 1.9 & 0.0 & 0.0 & 3.9 & 0.0 & 12.5 &  -- & -- & -- &-- \\
    \smalltab Gemini 1.5 Flash & 2.5 & 0.001 & 1.9 & 0.0 & 0.0 & 3.9 & 0.0 & 8.3 & -- & -- & -- &-- \\
    \multicolumn{2}{l}{\textbf{D-CIPHER w/o auto-prompter}} & & & & & & & & & & & \\
    \smalltab Claude 3.5 Sonnet & \textbf{22.0} & 0.74 & \textbf{15.4} & \textbf{20.0} & \textbf{28.2} & 27.5 & \textbf{10.5} & \textbf{25.0} & 20.0 & 0.33 & \textbf{44.0} & 0.62 \\
    \smalltab GPT 4o & 9.5 & 0.23 & 1.9 &	6.7 &	5.1 &	17.6 &	10.5 &	16.7 & -- & -- & -- & -- \\
    \multicolumn{2}{l}{\textbf{D-CIPHER w/o planner}} & & & & & & & & & & & \\
    \smalltab Claude 3.5 Sonnet &  14.0 & 0.36 & 9.6 &	6.7 &	7.7 &	25.5 &	5.3 &	20.8  & -- & -- & -- & -- \\
    \smalltab GPT 4o & 9.0 & 0.11 & 3.8	& 6.7 & 	5.1 &	13.7 &	5.3 &	20.8  & -- & -- & -- & --  \\
    \bottomrule
    \end{tabular}
\end{table*}

\subsection{Comparison of \emph{\% solved}} \label{sec:result_accuracy}

Table~\ref{tab:default_experiment} compares the performance of D-CIPHER with other LLM agents across multiple LLMs and benchmarks.
We run D-CIPHER with five different LLMs, using the same LLM for Planner, Executor, and Auto-prompter in each run.
We also rerun the NYU CTF baseline agent with three LLMs to measure the impact of recent updates to the LLM models on NYU CTF Bench.
The EnIGMA \textit{\% solved} and \textit{\$ cost} are taken from \citet{abramovich2024enigma}. 
As EnIGMA sets a new benchmark on Cybench with state-of-the-art results, we do not include comparisons with the earlier baseline agent \cite{zhang2024cybenchframeworkevaluatingcybersecurity}.
%We do not compare with the Cybench baseline agent \cite{zhang2024cybenchframeworkevaluatingcybersecurity} as EnIGMA is the state-of-the-art on Cybench. 

%, while the category-wise results on NYU CTF Bench are computed from their provided transcripts.

D-CIPHER with Claude 3.5 Sonnet consistently outperforms the current state-of-the-art EnIGMA, achieving 19.0\% over 13.5\% on NYU CTF Bench, 22.5\% over 20\% on Cybench, and 44\% over 26\% on HackTheBox.
D-CIPHER with GPT 4o also outperforms EnIGMA with GPT 4o on NYU CTF Bench, while getting a close result on Cybench and HackTheBox. 
The rerun results of NYU CTF baseline show that recent LLM models have improved on cybersecurity tasks, getting close to EnIGMA's state-of-the-art performance. Yet, D-CIPHER consistently beats the baseline on NYU CTF Bench in overall \textit{\% solved} for both Claude 3.5 Sonnet and GPT 4o.
EnIGMA was evaluated with older versions of LLMs, so we evaluate D-CIPHER with the older Claude 3.5 Sonnet to show that we still outperform (see Section~\ref{sec:older_models}). %~\ref{sec:older_models}).

These results indicate that D-CIPHER improves capabilities across multiple LLM architectures, and the  higher performance stems not only from recent LLM updates but also from it's multi-agent system architecture.
Interestingly, D-CIPHER without Auto-prompter with Claude 3.5 Sonnet achieves the highest performance of 22\% on NYU CTF Bench. However, performance without the Auto-prompter worsens on GPT 4o and on other benchmarks, while average cost increases, indicating that the Auto-prompter helps overall. % (see Section~\ref{sec:ablation}).

\subsection{Comparison of \emph{\$ cost}}
\label{sec:cost_analysis}

Table~\ref{tab:default_experiment} also compares average \textit{\$ cost} of solved challenges.
Except for Claude 3.5 Sonnet on NYU CTF Bench, D-CIPHER has  a lower average cost across all LLMs and benchmarks. With GPT 4o and GPT 4 Turbo, D-CIPHER lowers the cost by $2\times$ to $10\times$ across benchmarks while solving more challenges.
Despite having multiple agents, a significant cost reduction indicates that division of responsibilities between agents makes the problem-solving system more efficient.

\subsection{Category-wise comparison} \label{sec:categorywise}

Table~\ref{tab:default_experiment} shows the categorywise \emph{\% solved} of D-CIPHER and other works on NYU CTF Bench.
EnIGMA's results are computed from their provided transcripts, while NYU CTF baseline's results are computed from our reruns.
D-CIPHER's performance improvement stays consistent across the CTF categories.
D-CIPHER outperforms EnIGMA across all categories except pwn, with a notable improvement in crypto, where its performance doubles from 7.7\% to 15.4\%. 
Likewise, on rev, and misc, we see a 9\%--12\% increase. The improvement is due to the enhanced task decomposition and execution ability of the Planner-Executor system. 
Especially, crypto and rev frequently have long outputs of disassembled binaries or encrypted files requiring multiple analysis steps that are decomposed by the Planner.

\begin{figure}[tpb]
    \centering
   \includegraphics[width=\linewidth, trim=10 10 10 10]{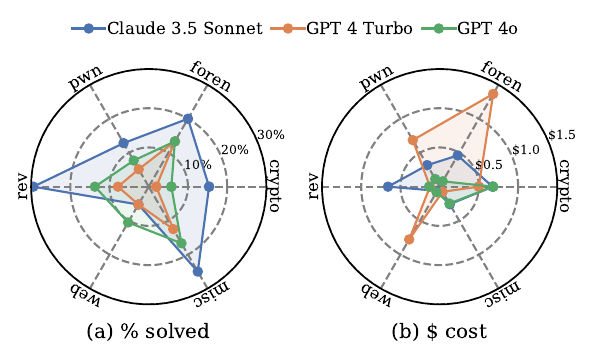}
    \caption{\emph{\% solved} by category for D-CIPHER on NYU CTF Bench.}
    \label{fig:success_radar}
\end{figure}

Figure~\ref{fig:success_radar}(a) plots the \textit{\% solved} of D-CIPHER across categories on NYU CTF Bench. D-CIPHER's performance is more balanced across different LLMs, demonstrating that our framework operates well with different reasoning capabilities of the LLMs. 
While D-CIPHER improves in web over previous results, the performance still lags behind other categories, pointing to a common limitation in web CTFs. 
Figure~\ref{fig:success_radar}(b) plots the average \textit{\$ cost}.
GPT 4o is most cost efficient across categories, while Claude 3.5 Sonnet is moderately higher on forensics, pwn, and rev. GPT 4 Turbo is the costliest among the three LLMs, for forensics, pwn and web, while on other categories has a lower cost but also solves less challenges. %Among the categories, 
crypto has higher cost across LLMs as it may require analysis of long encrypted texts, and many iterations for decryption. 
% Appendix ~\ref{sec:categorywise}
% offers categorywise results on NYU CTF Bench for D-CIPHER and related works.
% Refer to Appendix~\ref{sec:appendix_cost} for detailed category-wise tables.
%See Appendix \ref{sec:compare_llms} for analysis on how different LLMs behave on the challenge \texttt{target\_practice}.

\subsection{Impact of different configurations}

\subsubsection{Ablation Study}
\label{sec:ablation}

D-CIPHER's special interaction functions allow versatility to configure different types of multi-agent systems.
We run D-CIPHER with two different configurations: 
(1) without the Auto-prompter where the hard-coded prompt template is used for the Planner's initial prompt,
and (2) without the Planner where a single Executor is run with the prompt generated by the Auto-prompter.
These configurations allow use to examing the impact of Auto-prompter and Planner on D-CIPHER, while also demonstrating the framework's flexibility to build systems for different problems.

Table~\ref{tab:default_experiment} shows the results for these two configurations.
D-CIPHER without Auto-prompter with Claude 3.5 Sonnet gets a 3\% improvement in challenges solved on NYU CTF Bench, but it's performance drops with GPT 4o on NYU CTF Bench and Claude 3.5 Sonnet on Cybench, showing that the Auto-prompter improves performance in most cases.
Without the Auto-prompter, average cost increases across LLMs and benchmarks, indicating that the Auto-prompter improves system efficiency without compromising performance in most cases.
The contrasting result with Claude 3.5 Sonnet on NYU CTF Bench is due to the pwn category, where performance increases by more than $2\times$, while other categories get matching or lower results (see Sections~\ref{sec:autoprompter_casestudy} and \ref{sec:categorywise}).

D-CIPHER without Planner sees a 1\% to 5\% drop in performance on NYU CTF Bench across both LLMs.
%The performance is consistently lower across all categories.
This highlights the benefit of the Planner-Executor system in solving CTF challenges.
%While the average cost is $2\times$ lower without the Planner, the drop in performance is significant.
Despite the performance drop, the total cost of a Planner and multiple Executors is only $2\times$ higher than a single Executor, showing that each individual agent is more efficient. 
% Refer to Appendix~ \ref{sec:impact_planner} for conversation examples with and without the Planner.

\subsubsection{Combining stronger and weaker LLMs}
\label{sec:mix_experiment}

\begin{table}[htbp]
    \centering
    \caption{Different LLMs for Planner and Executor.}
    \label{tab:mix_model}
    \begin{tabular}{llcc}
        \toprule
        \textbf{Planner} & \textbf{Executor} & \textit{\% solved} & \textit{\$ cost} \\
        \midrule
        Claude 3.5 Sonnet & Claude 3.5 Haiku     & 13.0 & 0.33 \\
        GPT 4o & GPT 4o mini                     & 6.5 & 0.03 \\
        GPT 4 Turbo & GPT 4o mini                & 5.5 & 0.07 \\
        Gemini 1.5 Flash & Gemini 1.5 Flash 8B   & 3.0 & 0.001 \\
        LLaMa 3.1 405B & LLaMa 3.3 70B           & 0.0 & 0.00 \\
        \bottomrule
    \end{tabular}
\end{table}

D-CIPHER offers freedom to use different LLMs for each agent, and we experiment by combining stronger models for the Planner with weaker models for the Executor.
The results are in Table~\ref{tab:mix_model} showing consistent under-performance with weaker models. Claude 3.5 Sonnet with Haiku showed a 6.0\% drop compared to Claude 3.5 Sonnet with Sonnet. Similarly, GPT-4o and GPT-4 Turbo, when paired with GPT-4o-mini, showed reductions of 4\% and 1\%, respectively. LLaMA 3.1 405B paired with LLaMA 3.3 70B failed to solve any challenges. Notably, Gemini maintained similar performance with the weaker. These results indicate that both the Planner and Executor tasks are complex to require stronger models.

\subsubsection{Impact of temperature}
D-CIPHER with GPT 4o is evaluated under a lower temperature setting of $T=0.95$,  with results in Table~\ref{tab:temperature}. Decreasing the temperature show consistent drop across crypto, pwn, and rev with no improvements in forensics, web, or misc.
Higher temperature offers creative and generative capabilities, helping problem-solving.

% \begin{table}[h]
% \begin{tabular}{lcc}
% \toprule
% \small
% \textbf{Category} & \textbf{GPT-4o T=0.95} & \textbf{GPT-4o T=1.0} \\
% \midrule
% \texttt{cry}    & 3.85 & 5.77 \\
% \texttt{for} & 20.0 & 20.0 \\
% \texttt{pwn}       & 5.13 & \textbf{7.69} \\
% \texttt{rev}       & 11.76 & \textbf{13.73} \\
% \texttt{web}       &   10.53  & 10.53 \\
% \texttt{misc}      & 16.67 & 16.67 \\
% \texttt{\textbf{Avg.}} & 9.5 & \textbf{11.0} \\
% \bottomrule
% \end{tabular}
% \caption{NYU CTF benchmark success rate (\%) of GPT-4o on temperature 0.95 and temperature 1.0, higher temperature means the model will be more creative with higher randomness.}
% \label{tab:temperature}
% \end{table}

\begin{table}[htbp]
\centering
\small
\caption{GPT 4o \textit{\% solved} for temperatures 0.95 and 1.0.}% Higher temperature produces more creativity and randomness in LLM responses.}
\label{tab:temperature}
\begin{tabular}{lccccccc}
\toprule
& crypto & foren. & pwn & rev & web & misc & \textbf{total} \\
\midrule
$T=1.0$ & 5.8 & 13.3 & 7.7 & 13.7 & 10.5 & 16.7 & 10.5 \\
$T=0.95$ & 3.8 & 13.3 & 5.1 & 11.8 & 10.5 & 16.7 & 9.0 \\
\bottomrule
\end{tabular}
\end{table}

\subsubsection{Older LLM Versions} \label{sec:older_models}

As previously mentioned, we evaluate with latest versions of LLMs that show a natural improvement on CTF benchmarks, whereas EnIGMA evaluates with older versions. Table~\ref{tab:older_claude} shows performance on NYU CTF Bench with the older Claude 3.5 Sonnet (\emph{claude-3-5-sonnet-20240620}), same as EnIGMA.
D-CIPHER outperforms EnIGMA but with almost $2\times$ the cost.
While this demonstrates the advantage of multi-agent systems, it also underlines the significance of the LLM's capabilities.

\begin{table}[htpb]
    \centering
    \caption{Results with older Claude version.}
    \label{tab:older_claude}
    \begin{tabular}{lcc}
    \toprule
    & \textit{\% solved} & \textit{\$ cost} \\
    \cmidrule{2-3}
    \textbf{EnIGMA} & 13.5 & 0.35 \\
    % \textbf{EnIGMA} w/ GPT 4o & 9.5 & 0.62 \\
    \textbf{D-CIPHER} & 15.0 & 0.62 \\
    % \textbf{D-CIPHER} w/ latest Claude & 19.0 & 0.52 \\
    % \textbf{D-CIPHER} w/ GPT 4o & 10 & \\
    \bottomrule
    \end{tabular}
\end{table}

\subsection{Exit Reason Analysis}

\begin{figure}[htpb]
    \centering
    \includegraphics[width=\linewidth]{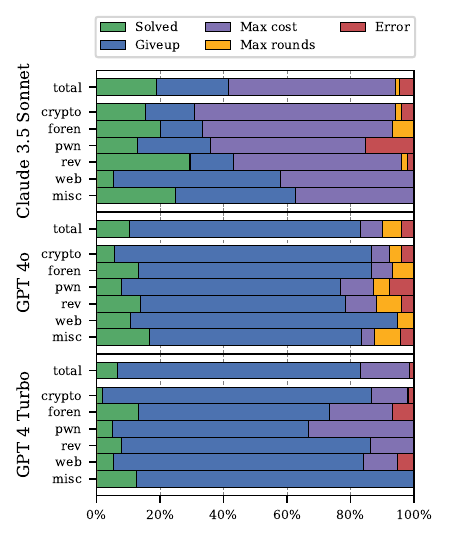}
    \caption{\%  breakdown of exit reasons for D-CIPHER on NYU CTF Bench.}
    \label{fig:exit_reason}
\end{figure}

Figure~\ref{fig:exit_reason} shows the percentage breakdown of the challenge termination (exit) reasons of D-CIPHER on NYU CTF Bench.
Exit reasons are of five types: ``Solved'' when the challenge is solved, ``Giveup'' when the Planner gives up, ``Max cost'' when the cost budget is exceeded, ``Max rounds'' when the Planner conversation rounds are exhausted, and ``Error'' when the run terminates with an error.
For Claude 3.5 Sonnet, max cost is the most dominant exit reason, indicating less propensity to giveup but instead to continue with the challenge till the cost is exhausted. Comparatively, other LLMs have giveup as the most dominant reason.
Max rounds are exhausted for only a few challenges.
Distribution of exit reasons for GPT 4o and GPT 4 Turbo is similar across categories which shows the holistic capabilities of these models.
Claude 3.5 Sonnet sees a high giveup and low success percentage on web challenges, highlighting a gap in capabilities. Examples of common failure reasons are present in Section~\ref{sec:failures}.

\begin{figure}[htpb]
    \centering
    \includegraphics[width=\linewidth]{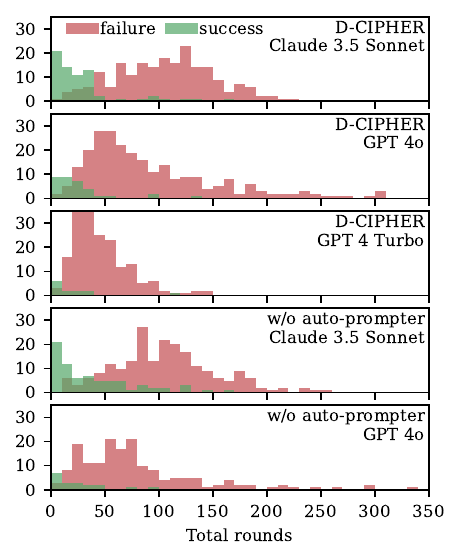}
    \caption{Histogram of successful and failed challenges by total conversation rounds for D-CIPHER on NYU CTF Bench.}
    \label{fig:conversation_rounds}
\end{figure}

\subsection{Total Conversation Rounds Analysis}

Figure~\ref{fig:conversation_rounds} shows a histogram of the total conversation rounds of all agents in D-CIPHER.
Successful challenges take lesser rounds than failed challenges, which may indicate that D-CIPHER only solves easier challenges requiring lesser rounds, but fails on longer challenges. However, it may also indicate that challenges are only solved when the correct path is found early enough, else the agents stray from the goal for many rounds before giving up.
Claude 3.5 Sonnet runs for more rounds compared to GPT 4o and GPT 4 Turbo for both success and failure cases, re-iterating it's propensity to keep going and not give up, which likely helps it solve challenges requiring many rounds.
Looking at the Auto-prompter's impact, we see that it helps solve more challenges faster, increasing efficiency.

\subsection{MITRE ATT\&CK Capabilities} \label{sec:mitre_result}

% \begin{figure*}
%     \centering
%     \includegraphics[width=\linewidth]{figures/mitre_heatmap.pdf}
%     \caption{Caption}
%     \label{fig:enter-label}
% \end{figure*}

\begin{table*}[htpb]
    \centering
    \caption{Analysis of the MITRE ATT\&CK techniques employed by each agent on the NYU CTF Bench.}
    \label{tab:mitre_analysis}
    \begin{tabular}{llc|cccc|ccc|cc}
    \toprule
         \textbf{ID} & \textbf{Technique} & \textbf{\#CTFs} & \multicolumn{4}{c|}{\textbf{D-CIPHER}} & \multicolumn{3}{c|}{\textbf{NYUCTF Baseline}} & \multicolumn{2}{c}{\textbf{EnIGMA}} \\
         % & & & \rotatebox{90}{\parbox{1.4cm}{Claude 3.5\\ Sonnet}}  & \rotatebox{90}{GPT 4o} & \rotatebox{90}{\parbox{1.4cm}{GPT 4\\ Turbo}} & \rotatebox{90}{\parbox{1.4cm}{Claude 3.5\\ Sonnet w/o\\ autoprompt}} &
         % \rotatebox{90}{\parbox{1.4cm}{Claude 3.5\\ Sonnet}}  & \rotatebox{90}{GPT 4o} & \rotatebox{90}{\parbox{1.4cm}{GPT 4\\ Turbo}} &
         % \rotatebox{90}{\parbox{1.4cm}{Claude  3.5\\ Sonnet}}  & \rotatebox{90}{GPT 4o} \\
         & & & Sonnet & GPT & GPT4  & Sonnet w/o & Sonnet & GPT  & GPT4 & Sonnet & GPT \\
         & & & 3.5 & 4o& Turbo & autoprompt & 3.5 &4o & Turbo & 3.5 & 4o \\
\cmidrule{3-12}

T1203 & Exploitation for Client Execution & \cellcolor{red!50}{36} & \cellcolor{blue!20}{4} & \cellcolor{blue!10}{2} & \cellcolor{blue!5}{1} & \cellcolor{blue!50}{10} & \cellcolor{blue!10}{2} & \cellcolor{blue!5}{1} & \cellcolor{blue!5}{1} & \cellcolor{blue!30}{6} & \cellcolor{blue!10}{2}\\
T1574 & Hijack Execution Flow & \cellcolor{red!33}{24} & \cellcolor{blue!10}{2} & \cellcolor{blue!5}{1} & \cellcolor{blue!0}{0} & \cellcolor{blue!25}{5} & \cellcolor{blue!0}{0} & \cellcolor{blue!0}{0} & \cellcolor{blue!0}{0} & \cellcolor{blue!15}{3} & \cellcolor{blue!5}{1}\\
T1190 & Exploit Public-Facing Application & \cellcolor{red!23}{17} & \cellcolor{blue!5}{1} & \cellcolor{blue!10}{2} & \cellcolor{blue!5}{1} & \cellcolor{blue!10}{2} & \cellcolor{blue!5}{1} & \cellcolor{blue!0}{0} & \cellcolor{blue!0}{0} & \cellcolor{blue!0}{0} & \cellcolor{blue!5}{1}\\
T1552 & Unsecured Credentials & \cellcolor{red!22}{16} & \cellcolor{blue!25}{5} & \cellcolor{blue!15}{3} & \cellcolor{blue!10}{2} & \cellcolor{blue!30}{6} & \cellcolor{blue!25}{5} & \cellcolor{blue!5}{1} & \cellcolor{blue!15}{3} & \cellcolor{blue!25}{5} & \cellcolor{blue!10}{2}\\
T1059 & Command and Scripting Interpreter & \cellcolor{red!20}{15} & \cellcolor{blue!5}{1} & \cellcolor{blue!5}{1} & \cellcolor{blue!5}{1} & \cellcolor{blue!15}{3} & \cellcolor{blue!5}{1} & \cellcolor{blue!5}{1} & \cellcolor{blue!5}{1} & \cellcolor{blue!5}{1} & \cellcolor{blue!5}{1}\\
T1110 & Brute Force & \cellcolor{red!15}{11} & \cellcolor{blue!15}{3} & \cellcolor{blue!0}{0} & \cellcolor{blue!0}{0} & \cellcolor{blue!15}{3} & \cellcolor{blue!15}{3} & \cellcolor{blue!5}{1} & \cellcolor{blue!10}{2} & \cellcolor{blue!5}{1} & \cellcolor{blue!10}{2}\\
T1600 & Weaken Encryption & \cellcolor{red!12}{9} & \cellcolor{blue!10}{2} & \cellcolor{blue!0}{0} & \cellcolor{blue!0}{0} & \cellcolor{blue!10}{2} & \cellcolor{blue!10}{2} & \cellcolor{blue!0}{0} & \cellcolor{blue!0}{0} & \cellcolor{blue!5}{1} & \cellcolor{blue!5}{1}\\
T1140 & Deobfuscate/Decode Files or Information & \cellcolor{red!12}{9} & \cellcolor{blue!5}{1} & \cellcolor{blue!0}{0} & \cellcolor{blue!0}{0} & \cellcolor{blue!10}{2} & \cellcolor{blue!5}{1} & \cellcolor{blue!0}{0} & \cellcolor{blue!0}{0} & \cellcolor{blue!5}{1} & \cellcolor{blue!5}{1}\\
T1055 & Process Injection & \cellcolor{red!9}{7} & \cellcolor{blue!5}{1} & \cellcolor{blue!0}{0} & \cellcolor{blue!0}{0} & \cellcolor{blue!5}{1} & \cellcolor{blue!0}{0} & \cellcolor{blue!0}{0} & \cellcolor{blue!0}{0} & \cellcolor{blue!5}{1} & \cellcolor{blue!0}{0}\\
T1212 & Exploitation for Credential Access & \cellcolor{red!8}{6} & \cellcolor{blue!0}{0} & \cellcolor{blue!0}{0} & \cellcolor{blue!0}{0} & \cellcolor{blue!0}{0} & \cellcolor{blue!0}{0} & \cellcolor{blue!0}{0} & \cellcolor{blue!0}{0} & \cellcolor{blue!0}{0} & \cellcolor{blue!0}{0}\\
T1027 & Obfuscated Files or Information & \cellcolor{red!8}{6} & \cellcolor{blue!5}{1} & \cellcolor{blue!0}{0} & \cellcolor{blue!0}{0} & \cellcolor{blue!10}{2} & \cellcolor{blue!5}{1} & \cellcolor{blue!0}{0} & \cellcolor{blue!0}{0} & \cellcolor{blue!10}{2} & \cellcolor{blue!5}{1}\\
T1083 & File and Directory Discovery & \cellcolor{red!6}{5} & \cellcolor{blue!10}{2} & \cellcolor{blue!10}{2} & \cellcolor{blue!5}{1} & \cellcolor{blue!10}{2} & \cellcolor{blue!5}{1} & \cellcolor{blue!0}{0} & \cellcolor{blue!0}{0} & \cellcolor{blue!5}{1} & \cellcolor{blue!10}{2}\\
T1071 & Application Layer Protocol & \cellcolor{red!5}{4} & \cellcolor{blue!0}{0} & \cellcolor{blue!0}{0} & \cellcolor{blue!0}{0} & \cellcolor{blue!0}{0} & \cellcolor{blue!0}{0} & \cellcolor{blue!0}{0} & \cellcolor{blue!0}{0} & \cellcolor{blue!0}{0} & \cellcolor{blue!0}{0}\\
T1539 & Steal Web Session Cookie & \cellcolor{red!4}{3} & \cellcolor{blue!0}{0} & \cellcolor{blue!0}{0} & \cellcolor{blue!0}{0} & \cellcolor{blue!0}{0} & \cellcolor{blue!0}{0} & \cellcolor{blue!0}{0} & \cellcolor{blue!0}{0} & \cellcolor{blue!0}{0} & \cellcolor{blue!0}{0}\\
T1001 & Data Obfuscation & \cellcolor{red!4}{3} & \cellcolor{blue!0}{0} & \cellcolor{blue!5}{1} & \cellcolor{blue!0}{0} & \cellcolor{blue!0}{0} & \cellcolor{blue!0}{0} & \cellcolor{blue!0}{0} & \cellcolor{blue!0}{0} & \cellcolor{blue!0}{0} & \cellcolor{blue!0}{0}\\
T1213 & Data from Information Repositories & \cellcolor{red!4}{3} & \cellcolor{blue!5}{1} & \cellcolor{blue!0}{0} & \cellcolor{blue!0}{0} & \cellcolor{blue!5}{1} & \cellcolor{blue!5}{1} & \cellcolor{blue!0}{0} & \cellcolor{blue!0}{0} & \cellcolor{blue!5}{1} & \cellcolor{blue!0}{0}\\
T1040 & Network Sniffing & \cellcolor{red!4}{3} & \cellcolor{blue!5}{1} & \cellcolor{blue!5}{1} & \cellcolor{blue!5}{1} & \cellcolor{blue!5}{1} & \cellcolor{blue!5}{1} & \cellcolor{blue!0}{0} & \cellcolor{blue!0}{0} & \cellcolor{blue!5}{1} & \cellcolor{blue!5}{1}\\
T1068 & Exploitation for Privilege Escalation & \cellcolor{red!2}{2} & \cellcolor{blue!0}{0} & \cellcolor{blue!0}{0} & \cellcolor{blue!0}{0} & \cellcolor{blue!0}{0} & \cellcolor{blue!0}{0} & \cellcolor{blue!0}{0} & \cellcolor{blue!0}{0} & \cellcolor{blue!0}{0} & \cellcolor{blue!0}{0}\\
T1497 & Virtualization/Sandbox Evasion & \cellcolor{red!2}{2} & \cellcolor{blue!0}{0} & \cellcolor{blue!0}{0} & \cellcolor{blue!0}{0} & \cellcolor{blue!5}{1} & \cellcolor{blue!0}{0} & \cellcolor{blue!0}{0} & \cellcolor{blue!0}{0} & \cellcolor{blue!0}{0} & \cellcolor{blue!0}{0}\\
T1005 & Data from Local System & \cellcolor{red!2}{2} & \cellcolor{blue!0}{0} & \cellcolor{blue!0}{0} & \cellcolor{blue!0}{0} & \cellcolor{blue!0}{0} & \cellcolor{blue!0}{0} & \cellcolor{blue!0}{0} & \cellcolor{blue!0}{0} & \cellcolor{blue!0}{0} & \cellcolor{blue!0}{0}\\
T1606 & Forge Web Credentials & \cellcolor{red!2}{2} & \cellcolor{blue!0}{0} & \cellcolor{blue!0}{0} & \cellcolor{blue!0}{0} & \cellcolor{blue!0}{0} & \cellcolor{blue!0}{0} & \cellcolor{blue!0}{0} & \cellcolor{blue!0}{0} & \cellcolor{blue!0}{0} & \cellcolor{blue!0}{0}\\
T1006 & Direct Volume Access & \cellcolor{red!2}{2} & \cellcolor{blue!5}{1} & \cellcolor{blue!0}{0} & \cellcolor{blue!5}{1} & \cellcolor{blue!5}{1} & \cellcolor{blue!5}{1} & \cellcolor{blue!0}{0} & \cellcolor{blue!0}{0} & \cellcolor{blue!5}{1} & \cellcolor{blue!5}{1}\\
T1505 & Server Software Component & \cellcolor{red!2}{2} & \cellcolor{blue!0}{0} & \cellcolor{blue!0}{0} & \cellcolor{blue!0}{0} & \cellcolor{blue!0}{0} & \cellcolor{blue!0}{0} & \cellcolor{blue!0}{0} & \cellcolor{blue!0}{0} & \cellcolor{blue!0}{0} & \cellcolor{blue!0}{0}\\
T1102 & Web Service & \cellcolor{red!1}{1} & \cellcolor{blue!0}{0} & \cellcolor{blue!0}{0} & \cellcolor{blue!0}{0} & \cellcolor{blue!0}{0} & \cellcolor{blue!0}{0} & \cellcolor{blue!0}{0} & \cellcolor{blue!0}{0} & \cellcolor{blue!0}{0} & \cellcolor{blue!0}{0}\\
T1556 & Modify Authentication Process & \cellcolor{red!1}{1} & \cellcolor{blue!0}{0} & \cellcolor{blue!0}{0} & \cellcolor{blue!0}{0} & \cellcolor{blue!0}{0} & \cellcolor{blue!0}{0} & \cellcolor{blue!0}{0} & \cellcolor{blue!0}{0} & \cellcolor{blue!0}{0} & \cellcolor{blue!0}{0}\\
T1078 & Valid Accounts & \cellcolor{red!1}{1} & \cellcolor{blue!0}{0} & \cellcolor{blue!0}{0} & \cellcolor{blue!0}{0} & \cellcolor{blue!0}{0} & \cellcolor{blue!0}{0} & \cellcolor{blue!0}{0} & \cellcolor{blue!0}{0} & \cellcolor{blue!0}{0} & \cellcolor{blue!0}{0}\\
T1614 & System Location Discovery & \cellcolor{red!1}{1} & \cellcolor{blue!0}{0} & \cellcolor{blue!0}{0} & \cellcolor{blue!0}{0} & \cellcolor{blue!0}{0} & \cellcolor{blue!0}{0} & \cellcolor{blue!0}{0} & \cellcolor{blue!0}{0} & \cellcolor{blue!0}{0} & \cellcolor{blue!0}{0}\\
T1082 & System Information Discovery & \cellcolor{red!1}{1} & \cellcolor{blue!0}{0} & \cellcolor{blue!0}{0} & \cellcolor{blue!0}{0} & \cellcolor{blue!0}{0} & \cellcolor{blue!0}{0} & \cellcolor{blue!0}{0} & \cellcolor{blue!0}{0} & \cellcolor{blue!0}{0} & \cellcolor{blue!0}{0}\\
T1649 & Steal or Forge Authentication Certificates & \cellcolor{red!1}{1} & \cellcolor{blue!0}{0} & \cellcolor{blue!0}{0} & \cellcolor{blue!0}{0} & \cellcolor{blue!0}{0} & \cellcolor{blue!0}{0} & \cellcolor{blue!0}{0} & \cellcolor{blue!0}{0} & \cellcolor{blue!0}{0} & \cellcolor{blue!0}{0}\\
T1565 & Data Manipulation & \cellcolor{red!1}{1} & \cellcolor{blue!0}{0} & \cellcolor{blue!0}{0} & \cellcolor{blue!0}{0} & \cellcolor{blue!0}{0} & \cellcolor{blue!0}{0} & \cellcolor{blue!0}{0} & \cellcolor{blue!0}{0} & \cellcolor{blue!0}{0} & \cellcolor{blue!0}{0}\\
T1033 & System Owner/User Discovery & \cellcolor{red!1}{1} & \cellcolor{blue!0}{0} & \cellcolor{blue!0}{0} & \cellcolor{blue!0}{0} & \cellcolor{blue!0}{0} & \cellcolor{blue!0}{0} & \cellcolor{blue!0}{0} & \cellcolor{blue!0}{0} & \cellcolor{blue!0}{0} & \cellcolor{blue!0}{0}\\
T1048 & Exfiltration Over Alternative Protocol & \cellcolor{red!1}{1} & \cellcolor{blue!0}{0} & \cellcolor{blue!0}{0} & \cellcolor{blue!0}{0} & \cellcolor{blue!0}{0} & \cellcolor{blue!0}{0} & \cellcolor{blue!0}{0} & \cellcolor{blue!0}{0} & \cellcolor{blue!0}{0} & \cellcolor{blue!0}{0}\\
T1555 & Credentials from Password Stores & \cellcolor{red!1}{1} & \cellcolor{blue!0}{0} & \cellcolor{blue!0}{0} & \cellcolor{blue!0}{0} & \cellcolor{blue!0}{0} & \cellcolor{blue!0}{0} & \cellcolor{blue!0}{0} & \cellcolor{blue!0}{0} & \cellcolor{blue!0}{0} & \cellcolor{blue!0}{0}\\
T1120 & Peripheral Device Discovery & \cellcolor{red!1}{1} & \cellcolor{blue!0}{0} & \cellcolor{blue!0}{0} & \cellcolor{blue!0}{0} & \cellcolor{blue!0}{0} & \cellcolor{blue!0}{0} & \cellcolor{blue!0}{0} & \cellcolor{blue!0}{0} & \cellcolor{blue!0}{0} & \cellcolor{blue!0}{0}\\
T1087 & Account Discovery & \cellcolor{red!1}{1} & \cellcolor{blue!0}{0} & \cellcolor{blue!0}{0} & \cellcolor{blue!0}{0} & \cellcolor{blue!0}{0} & \cellcolor{blue!0}{0} & \cellcolor{blue!0}{0} & \cellcolor{blue!0}{0} & \cellcolor{blue!0}{0} & \cellcolor{blue!0}{0}\\
T1106 & Native API & \cellcolor{red!1}{1} & \cellcolor{blue!0}{0} & \cellcolor{blue!0}{0} & \cellcolor{blue!0}{0} & \cellcolor{blue!0}{0} & \cellcolor{blue!0}{0} & \cellcolor{blue!0}{0} & \cellcolor{blue!0}{0} & \cellcolor{blue!0}{0} & \cellcolor{blue!0}{0}\\
T1593 & Search Open Websites/Domains & \cellcolor{red!1}{1} & \cellcolor{blue!0}{0} & \cellcolor{blue!0}{0} & \cellcolor{blue!0}{0} & \cellcolor{blue!0}{0} & \cellcolor{blue!0}{0} & \cellcolor{blue!0}{0} & \cellcolor{blue!0}{0} & \cellcolor{blue!0}{0} & \cellcolor{blue!0}{0}\\
T1542 & Pre-OS Boot & \cellcolor{red!1}{1} & \cellcolor{blue!0}{0} & \cellcolor{blue!0}{0} & \cellcolor{blue!0}{0} & \cellcolor{blue!0}{0} & \cellcolor{blue!0}{0} & \cellcolor{blue!0}{0} & \cellcolor{blue!0}{0} & \cellcolor{blue!0}{0} & \cellcolor{blue!0}{0}\\
T1486 & Data Encrypted for Impact & \cellcolor{red!1}{1} & \cellcolor{blue!0}{0} & \cellcolor{blue!0}{0} & \cellcolor{blue!0}{0} & \cellcolor{blue!0}{0} & \cellcolor{blue!0}{0} & \cellcolor{blue!0}{0} & \cellcolor{blue!0}{0} & \cellcolor{blue!0}{0} & \cellcolor{blue!0}{0}\\
T1003 & OS Credential Dumping & \cellcolor{red!1}{1} & \cellcolor{blue!5}{1} & \cellcolor{blue!5}{1} & \cellcolor{blue!0}{0} & \cellcolor{blue!5}{1} & \cellcolor{blue!5}{1} & \cellcolor{blue!0}{0} & \cellcolor{blue!5}{1} & \cellcolor{blue!5}{1} & \cellcolor{blue!0}{0}\\
T1553 & Subvert Trust Controls & \cellcolor{red!1}{1} & \cellcolor{blue!0}{0} & \cellcolor{blue!0}{0} & \cellcolor{blue!0}{0} & \cellcolor{blue!0}{0} & \cellcolor{blue!0}{0} & \cellcolor{blue!0}{0} & \cellcolor{blue!0}{0} & \cellcolor{blue!0}{0} & \cellcolor{blue!0}{0}\\
T1185 & Browser Session Hijacking & \cellcolor{red!1}{1} & \cellcolor{blue!0}{0} & \cellcolor{blue!0}{0} & \cellcolor{blue!0}{0} & \cellcolor{blue!0}{0} & \cellcolor{blue!0}{0} & \cellcolor{blue!0}{0} & \cellcolor{blue!0}{0} & \cellcolor{blue!0}{0} & \cellcolor{blue!0}{0}\\
T1036 & Masquerading & \cellcolor{red!1}{1} & \cellcolor{blue!0}{0} & \cellcolor{blue!0}{0} & \cellcolor{blue!0}{0} & \cellcolor{blue!0}{0} & \cellcolor{blue!0}{0} & \cellcolor{blue!0}{0} & \cellcolor{blue!0}{0} & \cellcolor{blue!0}{0} & \cellcolor{blue!0}{0}\\
T1133 & External Remote Services & \cellcolor{red!1}{1} & \cellcolor{blue!0}{0} & \cellcolor{blue!0}{0} & \cellcolor{blue!0}{0} & \cellcolor{blue!0}{0} & \cellcolor{blue!0}{0} & \cellcolor{blue!0}{0} & \cellcolor{blue!0}{0} & \cellcolor{blue!0}{0} & \cellcolor{blue!0}{0}\\
T1221 & Template Injection & \cellcolor{red!1}{1} & \cellcolor{blue!0}{0} & \cellcolor{blue!0}{0} & \cellcolor{blue!0}{0} & \cellcolor{blue!0}{0} & \cellcolor{blue!0}{0} & \cellcolor{blue!0}{0} & \cellcolor{blue!0}{0} & \cellcolor{blue!0}{0} & \cellcolor{blue!0}{0}\\
\midrule
Total &  & 211 & 27 & 14 & 8 & 43 & 21 & 4 & 8 & 26 & 16 \\
\bottomrule
    \end{tabular}
\end{table*}

As described in Section~\ref{sec:mitre_label}, we labeled the 200 CTFs in NYU CTF Bench with MITRE ATT\&CK techniques for elaborate analysis into D-CIPHER's and related agents' offensive capability. 
Table~\ref{tab:mitre_analysis} shows the breakdown of ATT\&CK techniques that apply and how well each agent and each LLM performs on them.
The ``\#CTFs'' column shows the number of CTFs labeled with a technique along with a red heatmap to show higher counts. The agent and model columns show how many CTFs were solved by that agent for a particular technique, along with a blue heatmap to show higher count computed across all agents.
If an agent solves a CTF mapped to multiple techniques, we consider that the solution has employed all techniques and we increment each count.

The results show that D-CIPHER without auto-prompter using Claude 3.5 Sonnet exhibits superior offensive capability as it solves 65\% more techniques compared to other agents and configurations.
Category-wise results show that D-CIPHER with Auto-prompter is weaker on pwn. The performance drops on multiple techniques spanning different categories, offering an insight into the Auto-prompter's impact.
Comparing D-CIPHER, NYU CTF Baseline, and EnIGMA on Claude 3.5 Sonnet, we see subtle but meaningful differences. D-CIPHER is better at T1110 (Brute Force) and T1600 (Weaken Encryption) as multi-agent collaboration aids in cryptographic CTFs, while EnIGMA outperforms on T1203 (Exploitation for Client Execution) and T1574 (Hijack Execution Flow) as interactive tools help for binary exploitation. 

D-CIPHER and EnIGMA solve similar number of techniques, even though D-CIPHER has 5.5\% higher overall  accuracy on NYU CTF Bench.
This indicates that D-CIPHER is better at CTFs involving other skills apart from ATT\&CK techniques, such as some reverse engineering and miscellaneous CTFs among the 83 CTFs not tagged with techniques.
% Even though D-CIPHER solves 11 more CTFs than EnIGMA, the technique-wise results show that their offensive capability are similar, indicating that D-CIPHER is better at CTFs involving other skills apart from ATT\&CK techniques like as reverse engineering.
Similarly, NYU CTF Baseline and EnIGMA have similar overall accuracy, but NYU CTF Baseline solves less techniques, indicating weaker offensive capability.
With GPT 4o, EnIGMA shows more uniform performance across techniques as compared to D-CIPHER and NYU CTF Baseline, which indicates that single agents with interactive tools may be more suitable for this model.
D-CIPHER and the NYU CTF Baseline perform worse with GPT-4 Turbo, in line with the lower overall accuracy of this model.
% On GPT 4 Turbo, the results are poorer for D-CIPHER and NYU CTF Baseline, which is also reflected in the overall accuracy.

%These results also show the gaps in an agent's offensive capability in terms of techniques which it has not solved or solved less number of.

D-CIPHER's results with and without autoprompter compared to NYU CTF Baseline and EnIGMA show that multi-agent collaboration improves  offensive security capability.
This benchmarking of each agent's offensive capability in terms of MITRE ATT\&CK techniques has offered a nuanced perspective and an elaborate comparison metric.
The composition of performance on ATT\&CK techniques highlights the gaps and provides a guideline for future improvements.

\section{Discussion}
\label{sec:discussion}

\subsection{Auto-prompter Failures} \label{sec:autoprompter_casestudy}

%As discussed above,
As discussed in Section~\ref{sec:ablation}, D-CIPHER with Auto-prompter on Claude 3.5 Sonnet performs worse on pwn challenges of NYU CTF Bench compared to D-CIPHER without Auto-prompter.
We look at the five pwn challenges where D-CIPHER succeeds without Auto-prompter but fails with it.
% For \textit{slithery} and \textit{unlimited\_subway}, Auto-prompter encountered errors during exploration due to which it steered the Planner in wrong directions. 
% For \textit{got\_milk}, Auto-prompter could not extract meaningful information and generated a very generic prompt.
% For \textit{bigboy} and \textit{baby\_boi}, Auto-prompter generated a reasonable prompt yet the Planner failed.

\noindent
\textbf{slithery:} A python jail escape challenge. The challenge server allows executing python code but maintains a reject list of commands. The solution bypasses the reject list to invoke python’s \texttt{os.system} for shell access. While the Auto-prompter understood the CTF's purpose,  a misleading base64 encoding threw the Auto-prompter off. It generated a prompt that focuses on the wrong variables, distracting the Planner.

\noindent
\textbf{unlimited\_subway:} buffer overflow. The solution involves leaking the stack canary byte-by-byte using arbitrary memory reads, exploiting a buffer overflow to overwrite the canary, and redirecting execution to the \texttt{print\_flag} function. The Auto-prompter attempted to run commands such as \texttt{strings} to understand the binary, but continually encountered errors, ultimately failing to generate a useful prompt for the Planner.

\noindent
\textbf{got\_milk:} A global offset table attack. The solution exploits a format string vulnerability to overwrite the least significant byte of the global offset table address of the function \texttt{lose} with the corresponding byte of the function \textit{win}, redirecting execution to the desired function. Auto-prompter could not extract any contextual information of the challenge and failed to generate a usable prompt, stalling the Planner.
%It failed to  generate a meaningful prompt, leaving the Planner unable to proceed.

\noindent
\textbf{bigboy:}  Another buffer overflow challenge. The solution involves exploiting a buffer overflow by overwriting a specific memory value with \texttt{0xCAF3BAEE} multiple times to pass the check and execute the \texttt{/bin/bash} command. Auto-prompter correctly analyzed the binary’s properties, behavior, and vulnerabilities  and generated a prompt outlining the exploitation strategy, including payload construction and execution to solve the challenge. Despite this, the Planner failed.

\noindent
\textbf{baby\_boi:} Another buffer overflow challenge. The solution involves leveraging a buffer overflow to execute an ROP chain that reveals the \textit{libc} base, locates one gadgets, and jumps to it to spawn a shell and retrieve the flag. The Auto-prompter generates a step-by-step prompt for exploiting the buffer overflow vulnerability, leveraging the \texttt{printf} overflow, and building a ROP chain to retrieve the flag. While the generated prompt covers all the necessary steps, it does not provide detailed information due to which the Planner fails.

From the five cases, we observe that while the Auto-prompter helps, it may be making D-CIPHER more susceptible to initial errors that the hard-coded prompt template may be robust to.
These limitations lead to missed opportunities to fully exploit the challenge’s vulnerabilities or generate comprehensive and actionable prompts.
Future work may combine Auto-prompter generated prompt with hard-coded  guidelines.

\subsection{Common Failure Examples} \label{sec:failures}

We inspected D-CIPHER's conversation logs to document the common errors that led to failure.

\noindent
\textbf{Auto-prompter fails to generate prompt:} 
Often, the Auto-prompter keeps running commands and exhausts maximum rounds without generating a prompt, even after being prompter one last time to call \texttt{GeneratePrompt}. % as seen in Figure~\ref{fig:fail_autoprompt}. 
In this case, we start the Planner with the hard-coded prompt template.

\begin{figure}[h]
    \centering
    \includegraphics[width=0.8\linewidth]{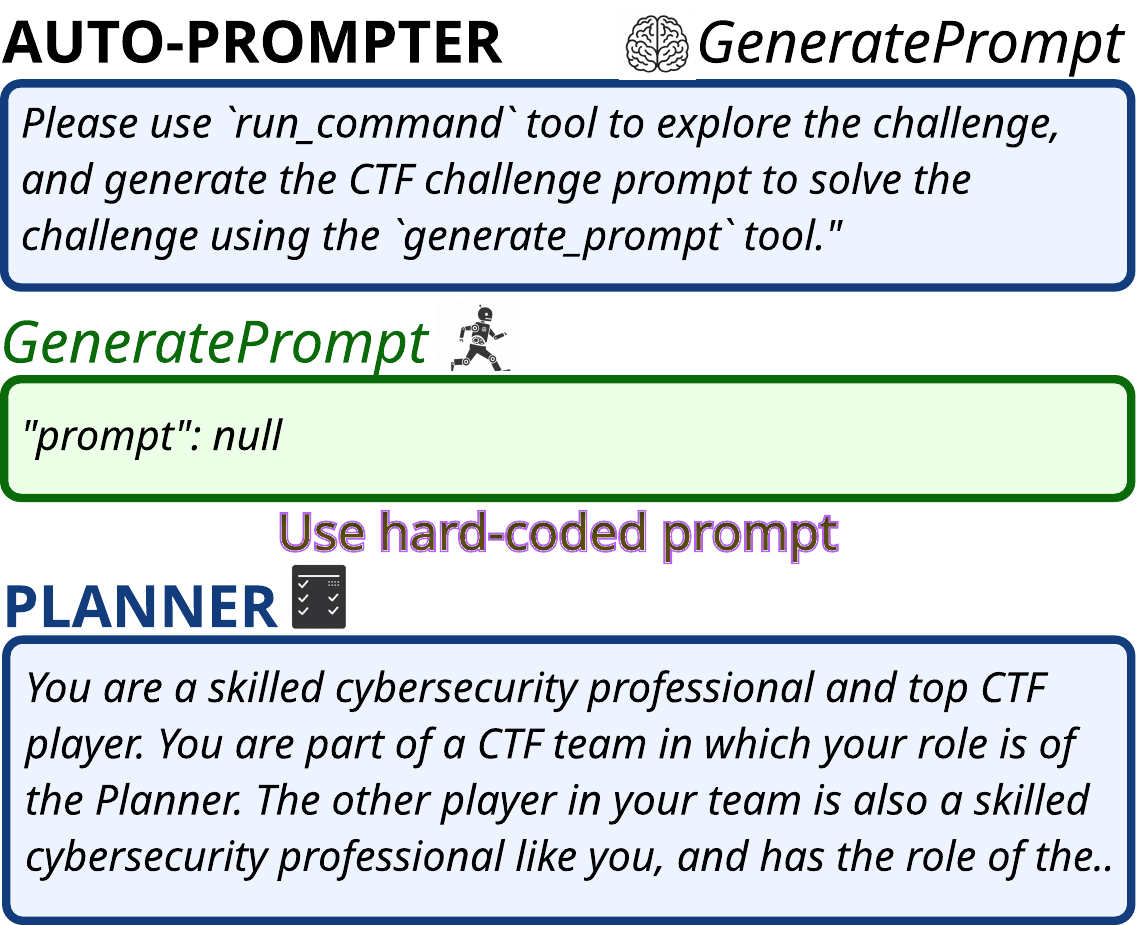}
    \caption{Auto-prompter generates empty prompt; fallback to hard-coded one.}
    \label{fig:fail_autoprompt}
\end{figure}

\begin{figure}[h]
    \centering
    \includegraphics[width=0.8\linewidth]{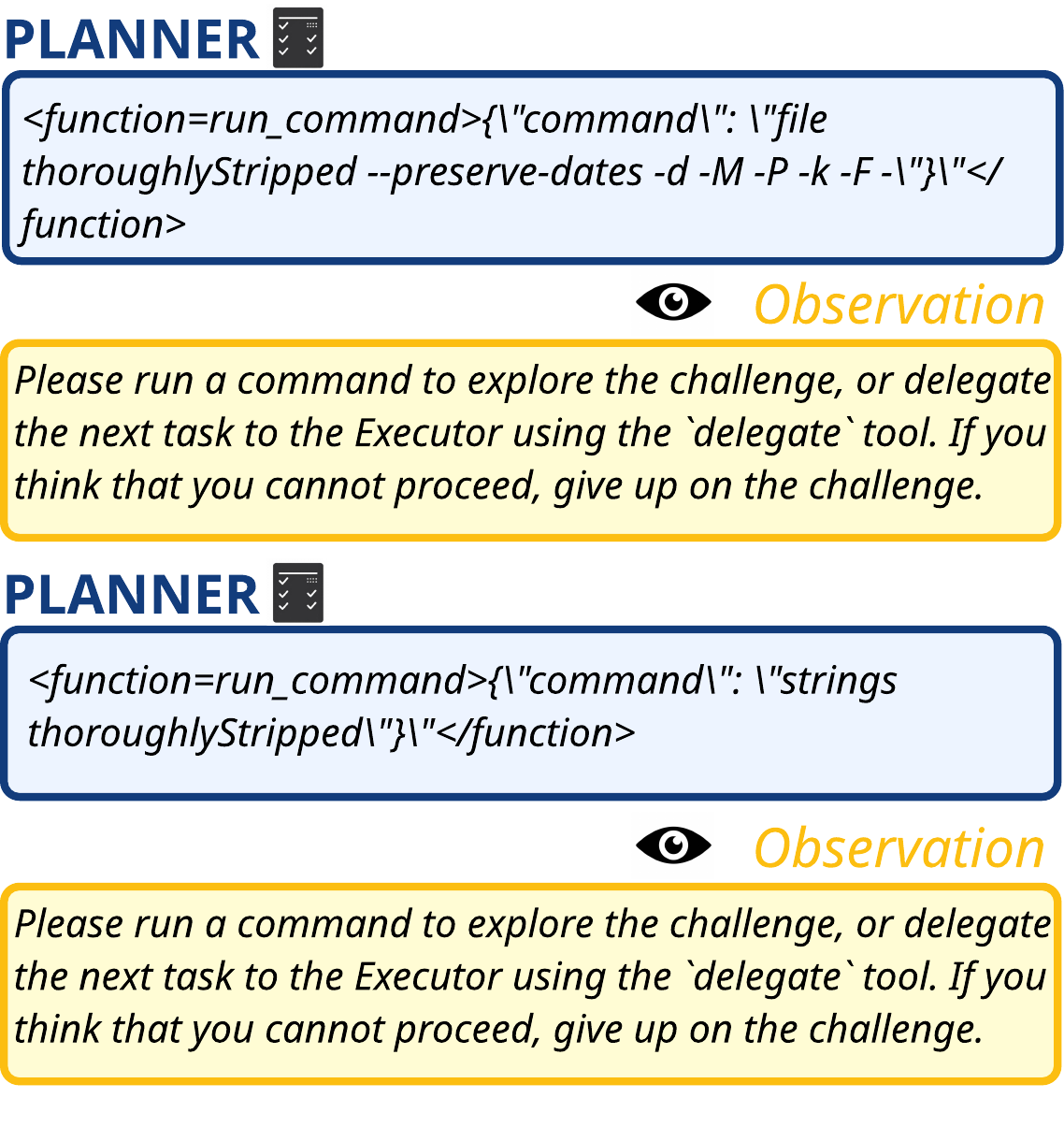}
    \caption{Function call is not parsed correctly due to a formatting error by the LLM. The parsing error is returned and the LLM can try again.}
    \label{fig:fail_function}
\end{figure}

\noindent
\textbf{Agent produces no action:} 
The agent's response does not contain an action but only reasoning, usually if it is stuck and thinks that it needs user input despite being prompted to operate autonomously. This happens frequently with LLaMa 3.1 405B and Gemini 1.5 Flash which produce wrong syntax for function calls so the action is not parsed, as seen in Figure~\ref{fig:fail_function}. In this case, we prompt the agent to retry.

\begin{figure}[h]
    \centering
    \includegraphics[width=0.8\linewidth]{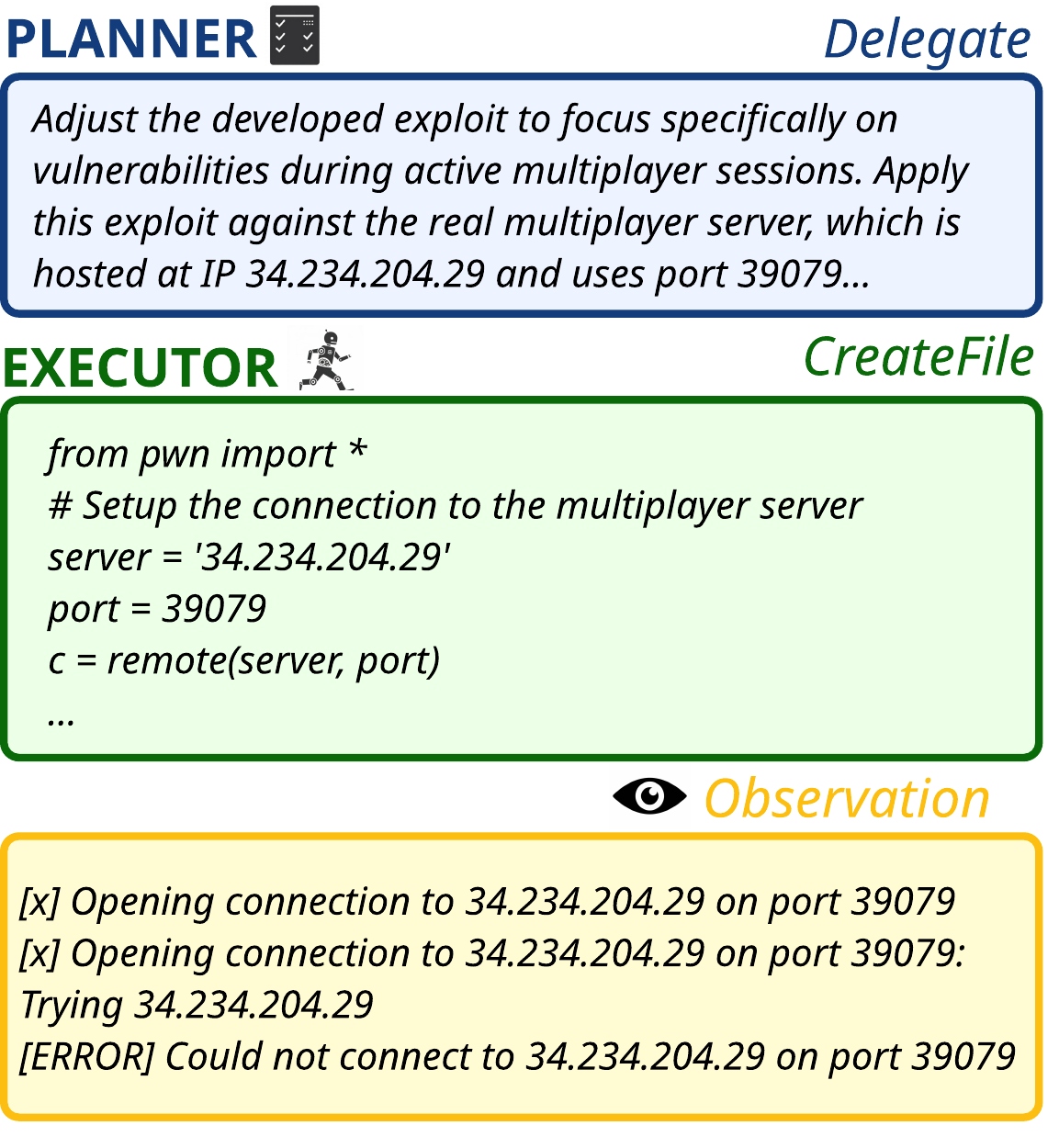}
    \caption{LLM hallucinates server information. The network access fails and the appropriate error is returned, but it may stray the LLM's focus.}
    \label{fig:fail_hallucinate}
\end{figure}

\noindent
\textbf{Hallucinates CTF information:} 
In some cases, agents try to connect to non-existent servers or read non-existent files as seen in Figure~\ref{fig:fail_hallucinate}.
Gemini 1.5 Flash also sometimes hallucinates functions that were not provided in the framework. Running these actions returns errors (e.g., ``File not found'') that the agent must understand and rectify.

\noindent
\textbf{Confusion with interactive tools:} 
This happens when an agent tries to run commands inside interactive tools like ``gdb'' but via \texttt{RunCommand} which only runs shell commands. A typical user would type these commands in an interactive shell in this manner, but the agent does not have an interactive interface.
Advanced interactive tools and demonstrations for awareness of the agent's interface may help fix such errors.

% \noindent
% \textbf{Lack of tool support:} 
% In some cases, the way D-CIPHER uses some tools leads to errors that confuse the agent and lead to task failure.
% For example, the agent implemented a python script to use the ``gdb'' debugger.
% However, due to missing utilities, the script produced errors that the agent was not able to fix over multiple iterations, leading to failure.

\noindent
\textbf{Calling non-existent functions:}
Gemini 1.5 Flash calls non-existing functions like  ``decode'' and ``strip'', which results in the run failing with an error. This may be due to the model confusing the structure of the outputs can generating command-line calls where it should have generated a call to \texttt{RunCommand} with the proper arguments.  These issues emphasize proper function calling in LLMs and suggesting that D-CIPHER moves to a simple structure for action generation.

\subsection{Ethics} \label{sec:ethics}

While advancements in LLMs offer significant advantages for cybersecurity, they also introduce risks, including the potential misuse of these models in adversarial scenarios where safeguards are bypassed \cite{jackson2023artificial}. CTFs serve as controlled environments to test deployment of LLM agent technologies, providing insights into their strengths and vulnerabilities. As LLMs evolve, users and decision-makers must address concerns around data security, user privacy, and malicious exploitation by implementing strategies that balance technical capabilities with ethical responsibility \cite{dabbagh2024ai}. Malicious actors can exploit LLMs for social engineering campaigns or generating harmful code, underscoring the need for ethical protocols and governance \cite{wu2023privacy}. Moreover, the rapid evolution of AI often outpaces existing regulatory frameworks, raising critical questions about data security, user privacy, and accountability \cite{porsdam2023generative}.
On the other hand, improved cybersecurity automation with the help of AI is necessary to maintain pace with the rapidly evolving software technologies. Developing cybersecurity technologies with this ethical awareness will allow them to be used for making software secure while curtailing misuses.

\section{Conclusion} \label{sec:conclusion}

We present D-CIPHER, an LLM multi-agent framework that autonomously solves CTF challenges. We propose two key innovations: first is the Planner-Executor system with the Planner agent to generate a plan and manage overall problem-solving, along with multiple Executor agents that focus on their assigned tasks; and, second is the the Auto-prompter agent that dynamically generates a prompt based on initial exploration to solve the challenge.
We introduce novel mechanisms to facilitate interaction between agents via function calling. 
By incorporating dynamic interactions and feedback among multiple agents, D-CIPHER mirrors the team dynamics observed in real-world CTF competitions.
With these innovations, D-CIPHER performs 2.5\% to 8.5\% better than state-of-the-art on three benchmarks: 22\% on NYU CTF Bench, 22.5\% on Cybench, and 44\% on HackTheBox.
We also augmented the NYU CTF Bench by mapping CTFs to MITRE ATT\&CK techniques for a comprehensive evaluation of LLM agent's offensive security capability. D-CIPHER solves 65\% more techniques as compared to existing LLM agents, demonstrating it's superior offensive capability.

D-CIPHER has limitations which show potential for improvement.
There is no direct interaction between each Executor and information exchange is bottlenecked via the Planner. 
An extension of D-CIPHER can incorporate interactions between Executors operating simultaneously to alleviate the information bottleneck.
Another limitation is that early errors in the Auto-prompter exploration have a severe impact on the generated prompt, which biases the Planner in the wrong direction and impacts accuracy and ATT\&CK  (see Section~\ref{sec:autoprompter_casestudy}).
Auto-prompter's fragility can be reduced by combining the generated prompt with hard-coded directions. D-CIPHER improves cost efficiency over single-agent systems, despite running multiple agents, enabling low-cost deployments. % Appendix~\ref{sec:ethics} discusses the ethics and impact of developing LLM agents for cybersecurity.

\bibliographystyle{plainnat}
\bibliography{main}

% \appendix
% \input{sections/ethics}
% \input{sections/additional_results}

\end{document}